\documentclass[numbers]{elsarticle}

\usepackage{graphicx}      
\usepackage{amsmath, amssymb}
\usepackage{epstopdf}
\usepackage{array}
\usepackage{float}
\usepackage{color}
\usepackage{url}
\usepackage{natbib}

\newcommand{\Real}{\mathbb{R}}
\newcommand{\Natural}{\mathbb{N}}
\newcommand{\Volume}{\mathcal{V}}
\newcommand{\Cov}{\mathrm{Cov}}
\newcommand{\bfx}{\boldsymbol{x}}
\newcommand{\bfw}{\boldsymbol{w}}
\newcommand{\bfy}{\boldsymbol{y}}
\newcommand{\bfz}{\boldsymbol{z}}

\newtheorem{defn}{Definition}
\newtheorem{thm}{Theorem}
\newtheorem{rem}{Remark}
\newtheorem{cor}{Corollary}
\newtheorem{lem}{Lemma}
\newtheorem{prop}{Proposition}

\definecolor{mycolor}{rgb}{0.7,0.3,0.3}

\begin{document}
\begin{frontmatter}

\title{One-Trial Correction of Legacy AI Systems and Stochastic Separation Theorems}

\author[Leic,NNov]{Alexander N. Gorban}\ead{ag153@le.ac.uk}
\author[Leic,ARM]{Richard Burton}\ead{Richard.Burton@arm.com}
\author[SpectralEdge]{Ilya Romanenko}\ead{ilya@spectraledge.co.uk}
\author[Leic,Leti,NNov]{Ivan Yu. Tyukin\corref{cor1}\fnref{f1}}\ead{I.Tyukin@le.ac.uk}

\address[Leic]{University of Leicester, Department of Mathematics, University Road, Leicester, LE1 7RH, United Kingdom}
\address[Leti]{Department of Automation and
        Control Processes, St. Petersburg State University of
        Electrical Engineering, Prof. Popova str. 5, Saint-Petersburg, 197376, Russian Federation}
\address[ARM]{Imaging and Vision Group, ARM Ltd, 1 Summerpool Rd, Loughborough, LE11 5RD, United Kingdom}
\address[SpectralEdge] {Spectral Edge Ltd, Bradfield Centre, 184 Science Park, Cambridge CB4 0GA, United Kingdom}
\address[NNov] {Lobachevsky State University of Nizhny Novgorod, Prospekt Ganarina 23, 603950,  Nizhny Novgorod, Russian Federation}

\cortext[cor1]{Corresponding author}
\fntext[f1]{The work was supported by the {Ministry of Education and Science} of Russia (Project No. 14.Y26.31.0022) and  Innovate UK  Knowledge Transfer Partnership grants KTP009890 and KTP010522.}

\begin{abstract}
We consider the problem of efficient ``on the fly'' tuning of existing, or {\it legacy}, Artificial Intelligence (AI) systems. The legacy AI systems are allowed to be of arbitrary class, albeit the data they are using for computing interim or final decision responses should posses an underlying structure of a high-dimensional topological real vector space. The tuning method that we propose enables dealing with errors without the need to re-train the system.  Instead of re-training a simple cascade of perceptron nodes is added to the legacy system. The added cascade modulates the AI legacy system's decisions. If applied repeatedly, the process results in a network of modulating rules ``dressing up'' and improving performance of existing AI systems.
Mathematical rationale behind the method is based on the fundamental property of measure concentration in high dimensional spaces. The method is illustrated with an example of fine-tuning a deep convolutional network that has been pre-trained to detect pedestrians in images.
\end{abstract}

\end{frontmatter}


\section{Introduction}

{\it Legacy} information  {\it systems}, i.e. information systems that have already been created and form crucial parts of existing solutions or service \cite{Legacy:1995}, \cite{Legacy:1999}, are common in engineering practice and scientific computing \cite{Kevrekidis}.  They are becoming particularly wide-spread, as legacy Artificial Intelligence  (AI) systems, in the areas of computer vision, image processing, data mining, and machine learning (see e.g. Caffe \cite{Caffe}, MXNet \cite{MXNet} and Deeplearning4j \cite{Deeplearning4j} packages).

Despite their success and utility, legacy AI systems  occasionally make mistakes.  Their generalization errors may be caused by a number of issues, for example, by incomplete, insufficient or obsolete information used in their development and design, or by oversensitivity to some signals, etc.  Adversarial images \cite{ICLR:2014:Szegedy}, \cite{Nguyen:2015} are well-known examples of such issues for classifiers based on Deep Learning Convolutional Neural Networks (CNNs) \cite{NIPS2012_4824}.  It has been shown in \cite{ICLR:2014:Szegedy} that imperceptible changes in the images used in the training set may cause labelling errors, and at the same time completely unrecognizable to human perception objects can be classified as the ones that have already been learnt \cite{Nguyen:2015}. Retraining such legacy systems requires resources, e.g. computational power, time and/or access to training sets, that are not always available. Thus solutions that reliably correct mistakes of legacy AI systems without retraining are needed.

%
%
%
%
%
%
%
%

Significant progress has been made to date to understand and mitigate atypical and spurious mistakes. For example, in \cite{Hansen:1990}, \cite{Ho:1995}, \cite{Ho:1998} using ensembles of classifiers was shown to improve performance of the overall system. With regards to adversarial images,  augmenting data  \cite{Kuznetsova:2015}, \cite{Misra:2015}, \cite{Prest:2012} and enforcing continuity of feature representations \cite{Zheng:2016} help to increase reliability of classification. These approaches nevertheless do not warrant error-free behavior; AI and machine learning systems drawing conclusions on the basis of empirical data are expected to make mistakes, as human experts occasionally do too \cite{Science:2016:FP}. Hence having a method for fast and reliable rectification of these errors in legacy systems is crucial.

In this work we present a computationally efficient solution to the problem of correcting AI systems' mistakes. In contrast to %
	conventional approaches \cite{Kuznetsova:2015}, \cite{Misra:2015}, \cite{Prest:2012}, \cite{Zheng:2016} focusing on altering data and improving design procedures of legacy AI systems, we propose that the legacy AI system itself be augmented by miniature low-cost and low-risk additions. These additions are, in their essence, small neural network cascades that in principle can be easily incorporated into existing architectures already employing neural networks as their inherent components. Importantly, we show that for a large class of legacy AI systems in which decision criteria are calculated on the basis of high-dimensional data representation, such small neuronal cascades can be constructed via simple non-iterative procedures. We prove this by showing that {\em in an essentially high-dimensional finite random set with probability close to one all points are extreme}, i.e. {\em every point is linearly separable from the rest of the set}. Such a separability holds even for large random sets up to some upper limit, which grows {\em exponentially} with dimension. This is proven for finite samples i.i.d. drawn from an equidistribution in a ball or ellipsoid and demonstrated also for equidistributions in a cube and for normal distribution. We can hypothesize now that this is a very general property of all sufficiently regular essentially high dimensional distributions.

Hence, if a legacy AI system employs high-dimensional internal data representation then a single element, i.e. a mistake, in this representation can be separated from the rest by a simple perceptron, for example, by the Fisher discriminant. Several such mistakes can be collated into a common corrector implementable as a small neuronal cascade. The results are based on ideas of measure concentration \cite{Gromov:1999}, \cite{GAFA:Gromov:2003}, \cite{Gorban:2007} and can be related to the works of  Gibbs \cite{Gibbs1902} and Levi \cite{Levi1951} and, on the other hand, to the Krein-Milman theorem {\cite{Krein-Milman:1940}}.

The paper is organized as follows. Section \ref{sec:problem} {  presents} formal statement of the problem, Section \ref{sec:main_results} contains mathematical background for developing One-Trial correctors of Legacy AI systems, in Section \ref{sec:discussion} {  we state and discuss practically relevant generalizations}, Section \ref{sec:examples} provides experimental results showing viability of our method for correcting state-of-the-art Deep Learning CNN classifiers, and Section \ref{sec:conclusion} concludes the paper. {  Main notational agreements are summarized in the Notation section below}.

\section*{Notation}

Throughout the paper the following notational agreements are used.
\begin{itemize}
\item $\Real$  denotes the field of real numbers;
\item $\Natural$  is the set of natural numbers;
\item $\Real^n$ stands for the $n$-dimensional real space; unless stated otherwise symbol $n$ is reserved to denote dimension of the underlying linear space;
\item let $\bfx\in\Real^n$, then $\|\bfx\|$ is the Euclidean norm of $\bfx$: $\|\bfx\|=\sqrt{x_1^2+\cdots + x_n^2}$;
\item {if $\bfx,\bfy$ are two elements of $\Real^n$ then $\langle \bfx,\bfy\rangle = \sum_{i=1}^n x_i y_i$;}
\item {symbol $\boldsymbol{0}_n$ denotes an element of $\Real^n$ of which the values of all of its $n$ coordinates are $0$; if the dimension $n$ is clear from the context, we will refer to such element as $\boldsymbol{0}$;}
\item $B_n(R)$ denotes a $n$-ball of radius $R$ centered at $\boldsymbol{0}_n$: $B_n(R)=\{\bfx\in\Real^n | \ \|\bfx\|\leq R\}$;
\item $\Volume(\Xi)$ is the Lebesgue volume of  $\Xi \subset \Real^n$;
\item $\mathcal{M}$ is an i.i.d. sample equidistributed in $B_n(1)$;
\item $M$ is the  number of points in $\mathcal{M}$, or simply the cardinality of the set $\mathcal{M}$.
\end{itemize}

{Let $E$ be a real vector space. Recall that a linear map $l:E\rightarrow\Real$ is called a linear functional \cite{Halmos:1974,Rudin:1991}. Similarly, an affine map $l:E\rightarrow\Real$, i.e. a map defined as $l(\bfx)=l_0(\bfx)+b$ where $l_0$ is a linear functional and $b\in\Real$,  is referred to as an affine functional.}

\section{Problem Formulation}\label{sec:problem}

Consider a legacy AI system that processes some {\it input} signals, produces {\it internal} representations of the input and returns some {\it outputs}. For the purposes of this work the exact nature and definition of the process and signals themselves are not important. Input signals may correspond to any physical measurements, outputs could be alarms or decisions, internal representations could include but not limited to extracted features, outputs of computational sub-routines etc. In the case of a CNN classifier inputs are images, internal signals are outputs of convolutional and dense layers, and outputs are labels of objects.

We assume, however, that some relevant information about the input, internal signals, and outputs can be combined into a common object, $\bfx$, representing, but not necessarily defining, the {\it state} of the AI system. The objects $\bfx$ are assumed to be elements of $\Real^n$. Over relevant  period of activity the AI system generates a set $\mathcal{M}$ of representations $\bfx$. Since we do not wish to impose any prior knowledge of how $\bfx$ are constructed {and following standard assumptions in machine learning literature \cite{Vapnik2000}}, it is natural to assume that $\mathcal{M}$ is a random sample drawn from some {distribution}.

\begin{figure}
\centering
\includegraphics[width=0.6\linewidth]{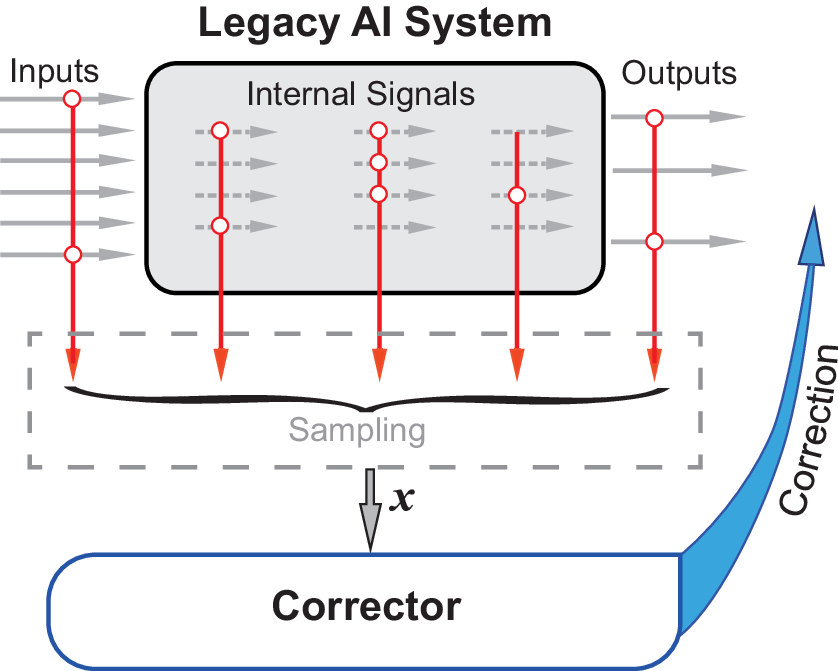}
\caption{Corrector of Legacy AI systems}\label{fig:corrector}
\end{figure}

We suppose that some elements of the set $\mathcal{M}$ are labelled as those corresponding to ``errors'' or mistakes which the original legacy AI system made. The task is to single out these errors and augment the AI systems' response by an additional device, {\it a corrector}. A diagram illustrating this setting is shown in Fig. \ref{fig:corrector}.
The corrector system must in turn satisfy a list of properties ensuring that its deployment is practically feasible
\begin{enumerate}
 \item the elements comprising the corrector must be re-usable in the original AI system;
 \item the elements, as elementary learning machines, should be able to generalize;
 \item the elements must be efficient computationally;
 \item the corrector must allow for fast non-iterative learning;
 \item the corrector should not affect functionality of the AI system, for reasonably large  $\mathcal{M}$ with $|\mathcal{M}|\gg n$.
\end{enumerate}

For a broad range of AI systems, a candidate element satisfying requirements $1)-3)$ is the parameterized {  affine} functional
\begin{equation}\label{eq:linear_func}
l(\bfx)=\langle \bfx, \bfw\rangle + b, \ \bfw\in\Real^n, \ b\in\Real
\end{equation}
followed, if needed, by a suitable nonlinearity. It is  a major building block of neural networks (convolutional, deep and shallow) as well as in decision trees and support vector machines. It's computational efficiency  and generalization capabilities \cite{Vapnik:2000} are well-known too.  Whether remaining requirements could be guaranteed, however, is not clear. The problem therefore is to find an answer to this question.

In the next sections we show that, remarkably, in high dimensions the {  affine} functionals do have these properties, with high probability.

\section{Mathematical Background}\label{sec:main_results}

\subsection{Separation of single points in finite random sets in high dimensions}

Our basic example throughout this section is an equidistribution in the  unit ball\footnote{Partially, the ideas concerning equidistributions in $B_n(1)$ have been presented in a conference talk \cite{GorTyuRom2016}} $B_n(1)$ in $\Real^n$. { Genralizations to equidistribution in an ellipsoid will be formulated in Section \ref{sec:discussion.ellipsoid}, and other distributions will be discussed in Sections \ref{sec:discussion.other_distributions} and \ref{sec:examples}.}

\begin{defn} Let $X$ and $Y$ be subsets of $\Real^n$. We say that a linear functional $l$ on $\Real^n$ separates $X$ and $Y$ if there exists a $t\in\Real$ such that
\[
l(\bfx)> t > l(\bfy) \ \forall \ \bfx\in X, \ \bfy\in Y.
\]
\end{defn}

Let $\mathcal{M}$ be an i.i.d. sample drawn from the equidistribution on the  unit ball $B_n(1)$.
We begin with evaluating the probability that a single element $\bfx$ randomly and independently selected from the same equidistribution can be separated from $\mathcal{M}$ by a linear functional. This probability, denoted as $P_1(\mathcal{M},n)$, is estimated in the theorem below.

\begin{figure}
\begin{center}
\includegraphics[width=0.4\textwidth]{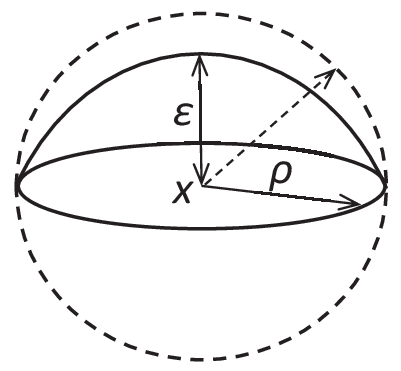}
\end{center}
\caption{A test point $\bfx$ and a spherical cap on distance $\varepsilon$ from the surface of unit sphere. Escribed sphere of radius $\rho$ is showed by dashed line. \label{fig:Segment}}
\end{figure}

\begin{thm}\label{theorem:point_separation}
Consider an equidistribution in a unit ball $B_n(1)$ in $\Real^n$, and let $\mathcal{M}$ be an i.i.d. sample from this distribution. Then
\begin{equation}\label{eq:random_point_separation}
\begin{split}
P_1(\mathcal{M},n)&\geq \max_{\varepsilon\in(0,1)} (1-(1-\varepsilon)^n)\left(1-\frac{{\rho}(\varepsilon)^n}{2}\right)^M,\\
{\rho}(\varepsilon)&=(1-(1-\varepsilon)^2)^{\frac{1}{2}}
\end{split}
\end{equation}
\end{thm}
{\it Proof of Theorem \ref{theorem:point_separation}.} The proof of the theorem is contained mostly in the following lemma
\begin{lem}\label{lem:given_point_separation} Let $y$ be random point from an equidistribution on a unit ball $B_n(1)$. Let $\bfx\in B_n(1)$ be a point inside the ball with $1>\|\bfx\|>1-\varepsilon>0$. Then
\begin{equation}\label{eq:given_point_separation}
P\left( \left\langle \frac{\bfx}{\|\bfx\|}, \bfy\right\rangle <1-\varepsilon\right) \geq 1-\frac{{\rho}(\varepsilon)^n}{2}.
\end{equation}
\end{lem}
{\it Proof of Lemma \ref{lem:given_point_separation}.} Recall that \cite{Levi1951}:
$\Volume(B_n(r))=r^n \Volume(B_n(1))$ for all $n\in\Natural$ $r>0$. The point $\bfx$ is inside the spherical cap $C_n(\varepsilon)$:
\begin{equation}\label{eq:spherical_cap}
C_n(\varepsilon)= B_n(1)\cap \left\{\boldsymbol{\xi}\in\Real^n \left|  \  \left\langle \frac{\bfx}{\|\bfx\|}, \boldsymbol{\xi} \right\rangle  > 1-\varepsilon \right\} \right..
\end{equation}
The volume of this cap can be estimated from above \cite{ball1997elementary} (see Fig. \ref{fig:Segment}) as
\begin{equation}\label{eq:cap_volume_estimate}
\Volume (C_n(\varepsilon))\leq \frac{1}{2}\Volume(B_{n}(1))\rho(\varepsilon)^{n}.
\end{equation}
The probability that the point $\bfy\in\mathcal{M}$ is outside of $C_n(\varepsilon)$ is equal to $1-\Volume(C_n(\varepsilon))/\Volume(B_n(1))$. Estimate (\ref{eq:given_point_separation}) now immediately follows from (\ref{eq:cap_volume_estimate}). $\square$

Let us now return to the proof of the theorem.  If $\bfx$ is selected independently from the equidistribution on $B_n(1)$ then the probabilities that $\bfx=\boldsymbol{0}_n$ or that it is on the boundary of the ball are $0$. Let $\bfx\neq \boldsymbol{0}_n$ be in the interior of $B_n(1)$. According to Lemma \ref{lem:given_point_separation}, the probability that a linear functional $l$ separates $\bfx$ from a point $\bfy\in\mathcal{M}$ is larger than $1-\frac{1}{2} \rho(\varepsilon)^n$. Given that points of the set $\mathcal{M}$ are i.i.d. in accordance to the equidistribution on $B_n(1)$, the probability that $l$ separates $\bfx$ from $\mathcal{M}$ is no smaller than $(1-\frac{1}{2} \rho(\varepsilon)^n)^{M}$.

On the other hand
\[
P(1>\|\bfx\|>1-\varepsilon \, |  \, \bfx\in B_n(1))= (1-(1-\varepsilon)^n).
\]
Given that $\bfx$ and $\bfy\in\mathcal{M}$ are independently drawn from the same equidistribution and that the probabilities of randomly selecting the point $\bfx$ exactly on the boundary of $B_n(1)$ or in its centre are zero, we can conclude that
\begin{equation}\label{eq:single_neuron_separation_at_epsilon}
P_1(\mathcal{M},n)\geq (1-(1-\varepsilon)^n) \left(1-\frac{1}{2} \rho(\varepsilon)^n\right)^{M}.
\end{equation}
Finally, noticing that  (\ref{eq:single_neuron_separation_at_epsilon}) holds for all $\varepsilon\in(0,1)$ including the value of $\varepsilon$ maximizing the rhs of (\ref{eq:single_neuron_separation_at_epsilon}), we can conclude that (\ref{eq:random_point_separation}) holds true too. $\square$

\begin{rem} \normalfont For $\rho(\varepsilon)^n$  small (i.e. $\rho(\varepsilon)^n \ll 1$) the term $\left(1-\frac{\rho(\varepsilon)^n}{2}\right)^M$ can be approximated {by the exponential
$e^{-M\frac{\rho(\varepsilon)^n}{2}}$. In this approximation, the rhs of}  estimate (\ref{eq:random_point_separation}) becomes
\begin{equation}\label{eq:single_neuron_separation_apprrox}
\max_{\varepsilon\in(0,1)} (1-(1-\varepsilon)^n) e^{-M\frac{\rho(\varepsilon)^n}{2}}, \ \rho(\varepsilon)^n \ll 1,
\end{equation}
\end{rem}
{resulting in a convenient approximate lower bound for the probability $P_1(\mathcal{M},n)$}.

To see how large the probability $P_1(\mathcal{M},n)$ could become for already rather modest dimensions, e.g. for $n=50$, we estimate {the value of $P_1(M,50)$ invoking (\ref{eq:single_neuron_separation_apprrox}) and letting $\varepsilon=1/5$ and $\rho=3/5$. The corresponding approximate lower bound for $P_1(\mathcal{M},50)$ is}
\[
{0.999985727} \exp(-4\times 10^{-12}M).
\]
For $M\leq 10^9$ this figure is {bounded from below by $0.995952534$. We note that the original expression in the rhs of (\ref{eq:random_point_separation}) for the same values of $\varepsilon, n$ and $M$ results in  $0.995952506$ which is different from the approximation only in the $8$-th digit.} Thus in dimension $50$ (and higher) a random point is linearly separable from a random set of $10^9$ points ({drawn independently from the same equidistribution in $B_n(1)$}) with probability of {approximately} $0.996$.

{ \begin{rem} \normalfont If $\bfx$ is an element from the sample $\mathcal{M}$ then the probability that {the element} $\bfx$ is linearly separable from all other points in the sample {may be} bounded from below by {the values} of $P_1(\mathcal{M},n)$, {albeit with a possible replacement of $M$ with $M-1$ in (\ref{eq:random_point_separation}) for a higher-accuracy estimate}.
\end{rem}}

\begin{rem}\label{rem:ExpSample} \normalfont Let $\bfx\in\mathcal{M}$ be a given query point. This query point determines the value of $\varepsilon=1-\|\bfx\|$ as the least $\varepsilon$-thickening of the unit sphere containing $\bfx$. With probability $1$ the values of $\varepsilon$ belong to the open interval $(0,1)$. Let $p\in(0,1)$ be the desired probability  that $\bfx$ is separated from the rest of the sample $\mathcal{M}$. It is clear that the estimate $P_1(\mathcal{M},n)\geq p$ holds for $M$ from some interval $[1,\overline{M}]$. Interestingly, for $n$ large enough, the maximal number $\overline{M}$ is exponentially large in dimension $n$. Indeed, let us fix the values of $\varepsilon\in(0,1)$ and $p\in(0,1)$. Then we find the estimate of the maximal possible sample size for which $P_1(\mathcal{M},n)\geq p$ remains valid:
\[
\max \{\overline{M}\}\geq \frac{\ln (p)}{\ln\left(1-\frac{\rho(\varepsilon)^n}{2} \right)}-\frac{\ln(1-(1-\varepsilon)^n)}{\ln\left(1-\frac{\rho(\varepsilon)^n}{2}\right)}
\]
Using
\[
\frac{x}{x-1}\leq \ln(1-x)\leq - x 
\]
we conclude that
\[
\max \{\overline{M}\}\geq \left(\frac{1}{\rho(\varepsilon)}\right)^n C(n,\varepsilon)\\
\]
where
\[
C(n,\varepsilon)=2 \left(|\ln(p)|\left(1-\frac{\rho(\varepsilon)^n}{2}\right) - |\ln(1-(1-\varepsilon)^n)|\right).
\]
Observe that for any fixed $\varepsilon\in(0,1)$ there is an $N(\varepsilon)$ large enough such that $C(n,\varepsilon)\geq |\ln(p)|$ for all $n\geq N(\varepsilon)$. Hence, for $n$ sufficiently large the following estimate holds:
\begin{equation}\label{eq:capacity}
\max \{\overline{M}\}\geq e^{ n \ln \left(\rho(\varepsilon)^{-1}\right)} |\ln(p)|.
\end{equation}
\end{rem}
Equation (\ref{eq:capacity}) can be viewed as a {\it separation capacity} estimate of linear functionals. This estimate links the level {  of} desired performance specified by  $p$, maximal size of the sample, $M$, and parameters of the data, $n$ and $\varepsilon$.

\subsection{Extreme points of a random finite set}

So far we have discussed the question of separability of a single random point $\bfx$, drawn from the equidistribution on $B_n(1)$, from a random i.i.d. sample $\mathcal{M}$ drawn from the same distribution. In practice, however, the data or a training set are given or fixed. It is thus important to know if the ``point'' linear separability property formulated in Theorem \ref{theorem:point_separation} persists (in one form or another) when the test point $\bfx$ belongs to the sample $\mathcal{M}$ itself. In particular, the question is if the probability $P_M(\mathcal{M},n)$ that each point  $\bfy\in \mathcal{M}$ is linearly separable from  $\mathcal{M}\backslash \{\bfy\}$ is close to $1$ in high dimensions? If such a property does hold then one could conclude that in high dimensions with probability close to $1$ all points of $\mathcal{M}$ are the vertices (extreme points) of the convex hull of $\mathcal{M}$ and none of $\bfy\in \mathcal{M}$ is a  convex combination of other points. The fact that this is indeed the case follows from Theorem \ref{theorem:all_points_separation}

\begin{thm}\label{theorem:all_points_separation} Consider an equidistribution in a unit ball $B_n(1)$ in $\Real^n$, and let $\mathcal{M}$ be an i.i.d. sample from this distribution.Then
\begin{equation}\label{eq:all_random_points_separation}
\begin{split}
&P_M(\mathcal{M},n)\geq \\
&\ \ \max_{\varepsilon\in(0,1)} \left[(1-(1-\varepsilon)^n)\left(1-(M-1)\frac{{\rho}(\varepsilon)^n}{2}\right)\right]^M.
\end{split}
\end{equation}
\end{thm}
{\it Proof of Theorem \ref{theorem:all_points_separation}}. Let $P:\mathcal{F}\rightarrow [0,1]$ be a probability measure and $A_i\in\mathcal{F}$, $i=1,\dots, M$. It is well-known that
\begin{equation}\label{eq:prob_subadditive}
P(A_1 \vee A_2 \vee \ldots \vee A_M)\leq \sum_{i=1}^M P(A_i)
\end{equation}
The probability that a test point $\bfy$ is in the $\varepsilon$-vicinity of the boundary of $B_n(1)$ is $1-(1-\varepsilon)^n$.
Fix $\bfy\in\mathcal{M}$  and  construct spherical caps $C_n(\varepsilon)$ for each element in $\mathcal{M}\backslash \{\bfy\}$ as specified by (\ref{eq:spherical_cap}) but with $\bfx$ replaced by the corresponding points from $\mathcal{M}\backslash \{\bfy\}$. According to (\ref{eq:prob_subadditive}) {  and Lemma \ref{lem:given_point_separation}}, the probability that $\bfy$ is in any of these caps is no larger than
$(M-1)\frac{\rho(\varepsilon)^n}{2}$. Hence the probability that a point $\bfy\in\mathcal{M}$ is separable from $\mathcal{M}\backslash \{\bfy\}$ is larger or equal to
$(1-(1-\varepsilon)^n)(1-(M-1)\frac{\rho(\varepsilon)^n}{2})$. Given that points of $\mathcal{M}$ are drawn independently and that there are exactly $M$ points in $\mathcal{M}$, the probability that every single point is linearly separable from the rest satisfies (\ref{eq:all_random_points_separation}).
$\square$

\begin{rem}\label{rem:SimpleEstMulti} \normalfont Note that employing (\ref{eq:prob_subadditive}) one can obtain another estimate of $P_M$:
\begin{equation}\label{eq:SimpleEstMulti}
P_M(\mathcal{M},n)\geq 1-M(1-P_1(\mathcal{M},n)).
\end{equation}
\end{rem}
We can utilise this estimate together with (\ref{eq:capacity}) and estimate the maximal size of the sample from below. Indeed, if we require that  $P_M(\mathcal{M},n)\geq q$ for some probability $q$, $0<q<1$, then it is sufficient that $P_1(\mathcal{M},n)>p$, where $1-p=\frac{1}{M}(1-q)$. Using in (\ref{eq:capacity}) $|\ln p| >1-p$, we get that $P_M(\mathcal{M},n)>q$ if $M\leq \tilde{M}$ for some maximal value $\tilde{M}$, that satisfies the inequality
$$\tilde{M}\geq  e^{ n \ln \left(\rho(\varepsilon)^{-1}\right)}\frac{1-q}{\tilde{M}}.$$
Immediately from this inequality we get the explicit exponential estimate of the maximum of $\tilde{M}$ from below:
\begin{equation}\label{eq:capacityExtrem}
\max\{\tilde{M}\}\geq e^{ \frac{1}{2}n \ln \left(\rho(\varepsilon)^{-1}\right)}\sqrt{1-q}
\end{equation}

{ Similarly} to the example discussed in the end of the previous section, let us evaluate the right-hand side of (\ref{eq:all_random_points_separation}) for some fixed values of $n$ and $M$. If $n=50$, $M=1000$, and $\varepsilon=1/5$, $\rho=3/5$ then this estimate gives: $P_{M}(\mathcal{M},50)>0.985$.

\subsection{Two-functional (two-neuron) separation in finite sets}

\begin{figure}
\begin{center}
\includegraphics[width=0.4\textwidth]{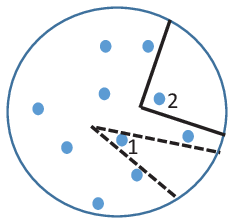}
\end{center}
\caption{Two-neuron separation in finite sets. Every point can be
separated by a sufficiently acute angle or highly correlated neurons:
Point 1 is separated from other points by an acute angle (dashed
lines). Point 2 is separated  by a right angle (non-correlated
neurons), but cannot be separated by a linear functional (i.e. by a
straight line).  \label{fig:Angles}}
\end{figure}

So far we have provided estimates of the probabilities that a single linear classifier or a learning machine can separate a given point from the rest of data and showed that two disjoint weakly compact subsets of a topological vector space can be separated by small networks of perceptrons. Let us now see how employing small networks may improve probabilities of separation of a point from the rest of the data in high dimensions. In particular, we will consider the case of a two-neuron separation in which the network is a simple cascade comprised of two perceptrons followed by a conjunction operation.

Before, however,  going any further we need to somewhat clarify and adjust the notion of separability of a point from a finite data set by a network so that the question makes any practical and theoretical sense. Consider, for instance, the problem of separating a test point by just two perceptrons. If one projects the data onto a $2$d plane so that projections of the test point and any other point from the rest of the data do not coincide then the problem always has a solution. This is illustrated with the diagram in Fig.~\ref{fig:Angles}. According to this diagram any given point from an arbitrary but finite data set could be cut out from the rest of the data by just two lines that intersect at a sufficiently acute angle. Thus two hyperplanes $\{\bfx\ | \ l_1(\bfx)=\theta_1\}$ and $\{\bfx\ | \ l_2(\bfx)=\theta_2\}$ whose projections onto the $2$d plane are exactly these two lines already constitute the two-neuron separating cascade.

The problem with this solution is that if  the acute angle determined by the inequalities  $l_1(\bfx)>\theta_1$, $l_2(\bfx)>\theta_2$  is small then the coefficients of the corresponding linear functionals $l_1(\bfx)$ and $l_2(\bfx)$, i.e. synaptic weights of the neurons, are highly correlated. Their Pearson correlation coefficient is close to $-1$. This implies that robustness of such a solution is low. Small changes in the coefficients of one perceptron could result in loss of separation. This motivates an alternative solution in which the coefficients are uncorrelated, i.e. the angles between two hyperplanes are right (or almost right).

Let us now analyse the problem of separation of a random i.i.d. finite sample $\mathcal{M}$ drawn from an equidistribution in $B_n(1)$ from a point $\bfx$ drawn independently from the same distribution by two non-correlated neurons. More formally, we are interested in the probability $\mathcal{P}_1(\mathcal{M},n)$ that a two-neuron cascade with uncorrelated synaptic weights separates $\bfx$ from $\mathcal{M}$. An estimate of this probability is provided in the next theorem.
\begin{thm}\label{theorem:two_neuron_separation} Consider an equidistribution in a unit ball $B_n(1)$ in $\Real^n$, and let $\mathcal{M}$ be an i.i.d. sample from this distribution. Then
\begin{equation}\label{eq:two_neuron_probability}
\begin{split}
\mathcal{P}_1(\mathcal{M},n)&\geq \max_{\varepsilon\in(0,1)} \ (1-(1-\varepsilon)^n)\times\\
&\left(1-\frac{\rho(\varepsilon)^n}{2}\right)^{M}\ e^{(M-n+1)\left[\frac{ \frac{\rho(\varepsilon)^n}{2}}{1-\frac{\rho(\varepsilon)^n}{2}}\right]}\times\\
&\left(1-\frac{1}{n!}\left((M-n+1) \frac{\frac{\rho(\varepsilon)^n}{2}}{1-\frac{\rho(\varepsilon)^n}{2}}\right)^n\right).
\end{split}
\end{equation}
\end{thm}
{\it Proof.} Observe that in the case of general position, a single neuron (viz. linear functional) separates $n+1$ points with probability $1$. This means that if no more than $n-1$ points from $\mathcal{M}$ are in the spherical cap $C_n(\varepsilon)$ corresponding to the test point then the second perceptron whose weights are orthogonal to the first one will filter out these additional spurious $n-1$ points with probability $1$.

Let $p_c$ be the probability that a point from $\mathcal{M}$ falls within the spherical cap $C_n(\varepsilon)$. Then the probability that only up to $n-1$ points of $\mathcal{M}$ will be in the cap $C_n(\varepsilon)$ is
\[
\mathcal{P}(M,n)= \sum_{k=0}^{n-1} \left(\begin{array}{c}
                                    M\\
                                    k
                                 \end{array}\right) (1-p_c)^{M-k} p_c^k
\]
Observe that  $\mathcal{P}(M,n)$, as a function of $p_c$,  is monotone and non-increasing on the interval $[0,1]$, with $\mathcal{P}(M,n)=0$  at $p_c=1$ and $\mathcal{P}(M,n)=1$ at $p_c=0$. Hence taking estimate (\ref{eq:cap_volume_estimate}) into account one can conclude:
\[
\mathcal{P}(M,n) \geq \sum_{k=0}^{n-1} \left(\begin{array}{c}
                                    M\\
                                    k
                                 \end{array}\right) \left(1-\frac{\rho(\varepsilon)^n}{2}\right)^{M-k} \left(\frac{\rho(\varepsilon)^n}{2}\right)^k.
\]

Noticing that
\[
\begin{split}
&\sum_{k=0}^{n-1} \left(\begin{array}{c}
                                    M\\
                                    k
                                 \end{array}\right) \left(1-\frac{\rho(\varepsilon)^n}{2}\right)^{M-k} \left(\frac{\rho(\varepsilon)^n}{2}\right)^k=\\
& \left(1-\frac{\rho(\varepsilon)^n}{2}\right)^M \sum_{k=0}^{n-1} \left(\begin{array}{c}
                                    M\\
                                    k
                                 \end{array}\right) \left(\frac{\frac{\rho(\varepsilon)^n}{2}}{1-\frac{\rho(\varepsilon)^n}{2}}\right)^k,
\end{split}
\]
and bounding $\left(\begin{array}{c}
                                    M\\
                                    k
                                 \end{array}\right)$
from above and below as
$\frac{(M-n+1)^k}{k!}\leq \left(\begin{array}{c}
                                    M\\
                                    k
                                 \end{array}\right)\leq \frac{M^k}{k!}$ for $ 0 \leq k\leq n-1$ we obtain
\[
\mathcal{P}(\mathcal{M},n)\geq \left(1-\frac{\rho(\varepsilon)^n}{2}\right)^M \sum_{k=0}^{n-1} \frac{1}{k!} \left(\frac{(M-n+1) \frac{\rho(\varepsilon)^n}{2} }{1-\frac{\rho(\varepsilon)^n}{2}}\right)^k.
\]
Invoking Taylor's expansion of $e^x$ at $x=0$ with the Lagrange reminder term:
\[
e^{x}=\sum_{k=0}^{n-1} \frac{x^k}{k!} + \frac{x^n}{n!}e^{\xi}, \ \xi\in [0,x],
\]
we can conclude that
\[
\sum_{k=0}^{n-1} \frac{x^k}{k!}\geq e^x\left(1-\frac{x^n}{n!}\right)
\]
for all $x\geq 0$. Hence
\[
\begin{split}
&\mathcal{P}(\mathcal{M},n)\geq\left(1-\frac{\rho(\varepsilon)^n}{2}\right)^M e^{(M-n+1)\left[\frac{ \frac{\rho(\varepsilon)^n}{2}}{1-\frac{\rho(\varepsilon)^n}{2}}\right]}\times\\
&\left(1-\frac{1}{n!}\left((M-n+1) \frac{\frac{\rho(\varepsilon)^n}{2}}{1-\frac{\rho(\varepsilon)^n}{2}}\right)^n\right).
\end{split}
\]
Finally, given that the probability that the test point $\bfx$ is in the $\varepsilon$-vicinity of the boundary of $B_n(1)$ is at least $(1-(1-\varepsilon)^n)$ and that $\bfx$ is independently selected form the same eqidistribution as the set $\mathcal{M}$, we obtain $\mathcal{P}_1(\mathcal{M},n)\geq (1-(1-\varepsilon)^n)\mathcal{P}(\mathcal{M},n)$. This in turn implies that (\ref{eq:two_neuron_probability}) holds. $\square$

\begin{figure}
\centering
\includegraphics[width=0.6\columnwidth]{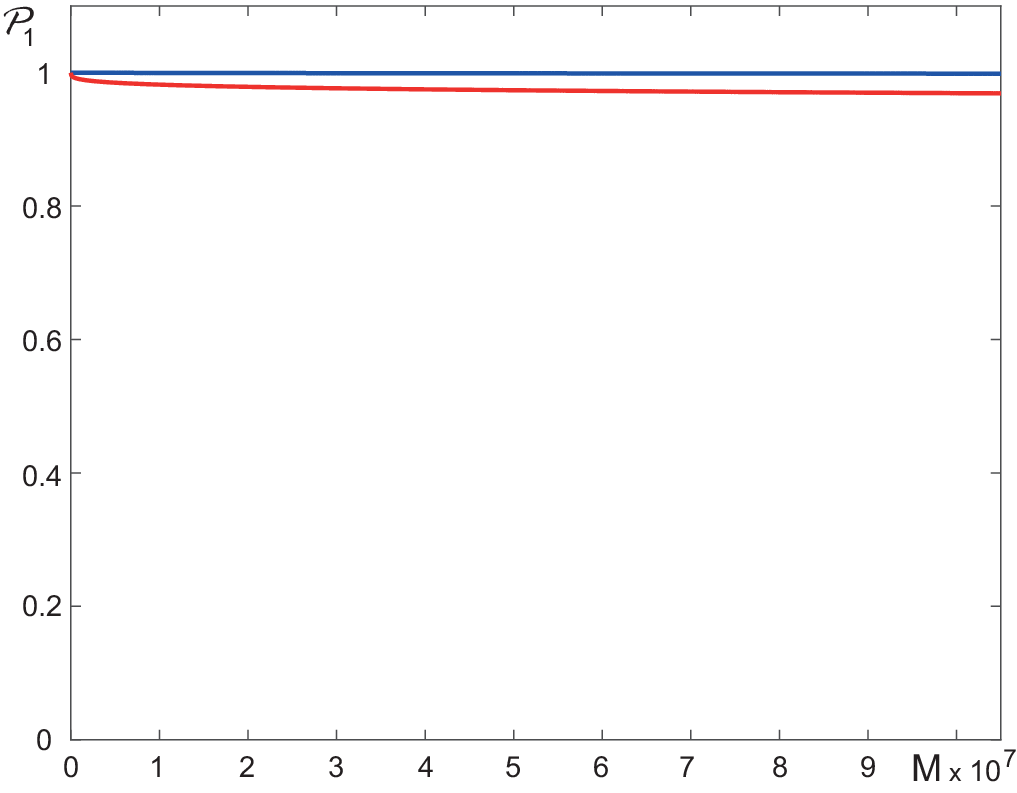}
\caption{Illustration to Theorem \ref{theorem:two_neuron_separation}. Blue line shows an estimate of the rhs of (\ref{eq:two_neuron_probability}) as a function of $M$ for $n=30$. Red line depicts an estimate of the rhs of (\ref{eq:random_point_separation}) as a function of $M$ for the same values of $n$.}\label{fig:two_neuron_separation_max}
\end{figure}

At the first glance estimate (\ref{eq:two_neuron_probability}) looks more complicated as compared to e.g. (\ref{eq:random_point_separation}). Yet, it differs from the latter by mere two factors. The first factor
\[
\left(1-\frac{1}{n!}\left((M-n+1) \frac{\frac{\rho(\varepsilon)^n}{2}}{1-\frac{\rho(\varepsilon)^n}{2}}\right)^n\right)
\]
is close to $1$ for $(M-n+1)\frac{\rho(\varepsilon)^n}{2} < 1$ and $n$ sufficiently large. The second factor:
\[
e^{(M-n+1)\left[\frac{ \frac{\rho(\varepsilon)^n}{2}}{1-\frac{\rho(\varepsilon)^n}{2}}\right]},
\]
is more important. It compensates for the decay of the probability of separation due to the term $(1-\frac{\rho(\varepsilon)^n}{2})^{M}$ keeping the rhs of (\ref{eq:two_neuron_probability}) close to $1$ over large interval of values of $M$. The effect is illustrated with
 Fig. \ref{fig:two_neuron_separation_max}. Observe that the probability corresponding to the two-neuron cascade remains in a close vicinity of $1$, and the probability of one-neuron separation gradually decays with $M$.

\begin{rem}\label{rem:give_two_neuron_separability} \normalfont Comparing performance of single vs two-neuron separability in terms of the probabilities $\mathcal{P}_1(\mathcal{M},n)$ involves taking the maximum of
\begin{equation}\label{eq:single_sepration_given_eps}
\mathcal{P}_1(\mathcal{M},n,\varepsilon)=(1-(1-\varepsilon)^n)\left(1-\frac{{\rho}(\varepsilon)^n}{2}\right)^M
\end{equation}
and
\begin{equation}\label{eq:two_neuron_sepration_given_eps}
\begin{split}
\mathcal{P}_1(\mathcal{M},n,\varepsilon)=&(1-(1-\varepsilon)^n)\times\\
&\left(1-\frac{\rho(\varepsilon)^n}{2}\right)^{M}\ e^{(M-n+1)\left[\frac{ \frac{\rho(\varepsilon)^n}{2}}{1-\frac{\rho(\varepsilon)^n}{2}}\right]}\times\\
&\left(1-\frac{1}{n!}\left((M-n+1) \frac{\frac{\rho(\varepsilon)^n}{2}}{1-\frac{\rho(\varepsilon)^n}{2}}\right)^n\right)
\end{split}
\end{equation}
with respect to $\varepsilon$ over $(0,1)$. In some situations, when the testing point is already given, the probabilities $\mathcal{P}_1(\mathcal{M},n)$ are no longer relevant since the value of $\varepsilon$ corresponding to the testing point is fixed. In this cases one needs to compare $\mathcal{P}_1(\mathcal{M},n,\varepsilon)$ defined by (\ref{eq:single_sepration_given_eps}) and (\ref{eq:two_neuron_sepration_given_eps}) instead. Performance of the corresponding separation schemes are illustrated with Fig. \ref{fig:two_neuron_separation}.  Notice that the two-neuron cascade significantly outperforms the single neuron one over a large interval of values of $M$.
\begin{figure}
\centering
\includegraphics[width=0.6\columnwidth]{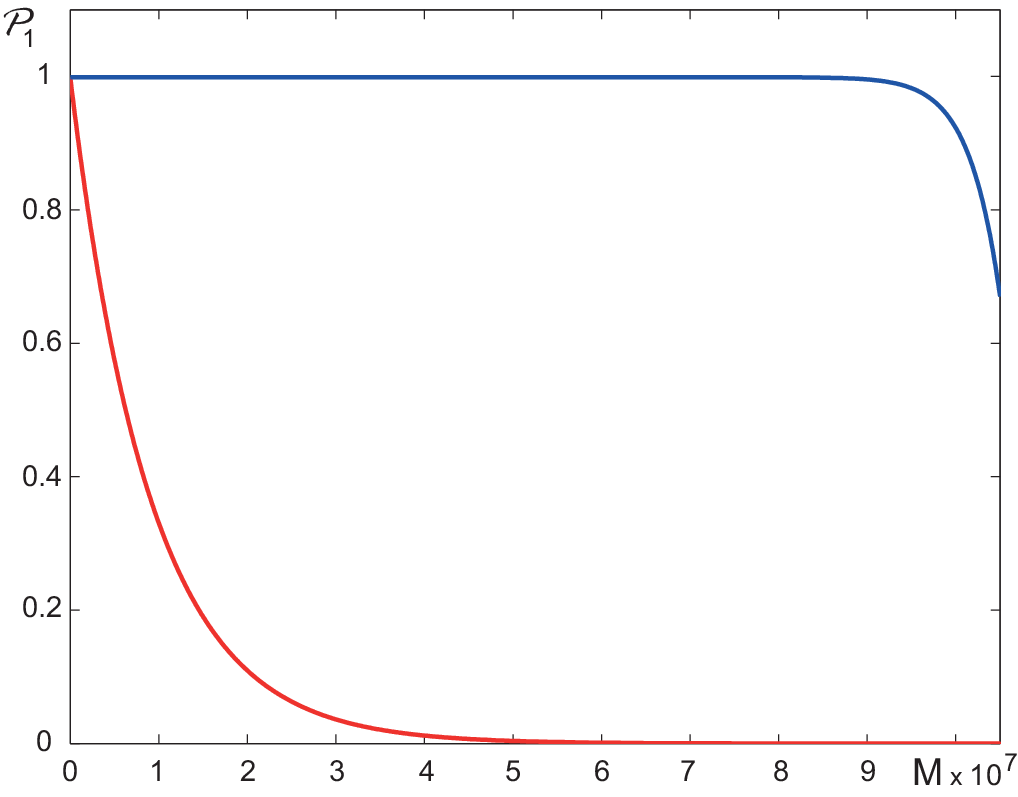}
\caption{Illustration to Remark \ref{rem:give_two_neuron_separability}. Blue line shows an estimate of the rhs of (\ref{eq:two_neuron_sepration_given_eps}) as a function of $M$ at $\varepsilon=1/5$, $\rho(\varepsilon)=3/5$, and $n=30$. Red line depicts an estimate of the rhs of (\ref{eq:single_sepration_given_eps}) as a function of $M$ for the same values of $\varepsilon$, $\rho(\varepsilon)$, and $n$.}\label{fig:two_neuron_separation}
\end{figure}

\end{rem}
\begin{rem} \normalfont
The probability $\mathcal{P}_M(\mathcal{M},n)$ that {\em each} point from $\mathcal{M}$ can be separated from other points by two uncorrelated neurons can be estimated like in Remark~\ref{rem:SimpleEstMulti}: $\mathcal{P}_M(\mathcal{M},n)\geq 1- M(1-\mathcal{P}_1(\mathcal{M},n))$.
\end{rem}

\subsection{Separation by small neural networks}\label{sec:small}


In the previous sections we analysed and discussed linear separability of single elements of $\mathcal{M}$. As far as the problem of a legacy AI system corrector is concerned, however, it is not difficult to envision a need to implement more complicated dependencies, including logical predicates or involving more complicated data geometry. Below we provide some notions and results enabling us to deal with these issues.

Consider a topological real vector space $L$. For every continuous linear functional $l$ on $L$ and a real number $\theta$ we define an open {\em elementary neural predicate} $l(\bfx)>\theta$ and a closed elementary predicate $l(\bfx)\geq \theta$ (`open' and `closed' are sets defined by predicates). The elementary neural predicate is true (=1) or false (=0) depending on whether the correspondent inequality holds. The negation of an open elementary neural predicate is a closed elementary neural predicate (and converse). A compound predicate is a Boolean expression composed of elementary predicates.

\begin{defn} A set $X \subset L$ is $k$-neuron separable from a set $Y \subset L$ if there exist $k$ elementary neural predicates $P_1,\ldots,P_k$ and a composed of them compound predicate, a Boolean function $B(\bfx)=B(P_1(\bfx),\ldots, P_k(\bfx))$, such that $B(\bfx)=$true for $\bfx\in X$ and $B(\bfx)=$false for $\bfx\in Y$.
\end{defn}

If $X$ consist of one point then the compound predicate in the definition of $k$-neuron separability can be selected in a form of conjunction of $k$ elementary predicates $B(\bfx)=P_1(\bfx)\&\ldots \& P_k(\bfx))$:

\begin{prop} Let $X=\{\bfz\}$ and $X$ is $k$-neuron separable from $Y \subset L$. Then there exist $k$ elementary neural predicates $P_1,\ldots,P_k$  such that the Boolean function $B(\bfx)=P_1(\bfx)\&\ldots \& P_k(\bfx)$ is true on $X$ and false on Y.
\end{prop}


{\it Proof}.  Assume that there exist $k$ elementary neural predicates $Q_1,\ldots,Q_k$ and a Boolean function $B(Q_1,\ldots,Q_k)$ such that $B$ is true on $\bfx$ and false on $Y$. Represent $B$ in a disjunctive normal form as a disjunction of conjunctive clauses:
\[
B=C_1\lor C_2 \lor \ldots \lor C_N,
\]
where each $C_i$ has a form  $C_i=R_1\& R_2 \& \ldots \& R_k$  and $R_i=Q_i$  or $R_i= \neg Q_i$.

At least one conjunctive clause $C_i(\bfx)$ is true because $B(\bfx)=$true. On the other hand, $C_i(\bfy)=$false for all $\bfy\in Y$ and $i=1,\ldots,N$ because their disjunction $B(\bfy)$=false. Let $C_i(\bfx)$=true. Then $C_i$ separates $\bfx$ from $Y$ and other conjunctive clauses are not necessary in $B$. We can take $B=C_i=R_1\& R_2 \& \ldots \& R_k$,  where $R_i=Q_i$  or $R_i= \neg Q_i$. Finally, we take $P_i=R_i$. $\square$


In contrast to Hahn-Banach theorems of linear separation, the $k$-neuron separation of sets does not require any sort of convexity. Weak compactness is sufficient. Assume that continuous linear functionals separate points in $L$. Let $Y\subset L$ be a weakly compact set \cite{Schaefer1999} and $\bfx\notin Y$.

\begin{prop}\label{Prop:PointFromSet}   $\bfx$ is $k$-neuron separable from $Y$ for some $k$.
\end{prop}


 {\it Proof.} For each $\bfy \in Y$ there exists a continuous linear functional $l_{\bfy}$ on $L$ such that $l_{\bfy}(\bfx)=l_{\bfy}(\bfy)+2$ because continuous linear functionals separate points in $L$. Inequality $l_{\bfy}(\bfz)<l_{\bfy}(\bfx)-1$ defines an open half-space $L_{\bfy}^<=\{\bfz\ | \ l_{\bfy}(\bfz)<l_{\bfy}(\bfy)+1\}$, and $\bfy\in L_{\bfy}^<$. The collection of sets $\{L_{\bfy}^< \ | \ \bfy \in Y\}$ forms an open cover of $Y$. Each set $L_{\bfy}^<$ is open in weak topology and $Y$ is weakly compact, hence there exists a final covering of $Y$ by sets $L_{\bfy}^<$:
\[
Y \subset \bigcup_{i=1,\ldots, k} L_{\bfy_i}^<
\]
for some finite subset $\{\bfy_1, \ldots, \bfy_k\} \subset Y$. The inequality $l_{\bfy_i}(\bfx)> l_{\bfy_i}(\bfy_i)+\alpha$ holds  for all $i=1, \ldots, k$ and $\alpha <2$. Let us select $1<\alpha<2$ and take elementary neural predicates $P_i^{\alpha}(\bfx)=(l_{\bfy_i}(\bfx)> l_{\bfy_i}(\bfy_i)+\alpha)$. The conjunction $P_1^{\alpha}\&\ldots \&P_k^{\alpha}$ is true on $\bfx$. Each point $\bfy\in Y$ belongs to $L_{\bfy_i}^<$ for at least one $i=1, \ldots, k$. For this $i$, $P_i^{\alpha}(\bfy)$=false. Therefore, $P_1^{\alpha}\&\ldots \&P_k^{\alpha}$ is false on $Y$. Hence, this conjunction separates $\bfx$ from $Y$. $\square$

\begin{rem} \normalfont
Note that we can select two numbers $1<\beta<\alpha<2$. Both $P_1^{\alpha}\&\ldots \&P_k^{\alpha}$ and $P_1^{\beta}\&\ldots \&P_k^{\beta}$ separate $\bfx$ from $Y$.
\end{rem}

The theorem about $k$-neuron separation of weakly compact sets follows from Proposition~\ref{Prop:PointFromSet}. Let $X$ and $Y$ be disjoint and weakly compact subsets of $L$, and continuous linear functionals separate points in $L$. Then the following statement holds

\begin{thm}   $X$ is $k$-neuron separable from $Y$ for some $k$.
\end{thm}

{\it Proof.}   For each $\bfx\in X$ apply the construction from the proof of Proposition~\ref{Prop:PointFromSet} and construct the conjunction $B_{\bfx} =P_1^{\alpha}\&\ldots \&P_k^{\alpha}$, which separates $\bfx$ from $Y$. Every set of the form $F_{\bfx}=\{\bfz\in L\ | \ B_{\bfx}(\bfz)=\mbox{true}\}$ is intersection of a finite number of open half-spaces and, therefore, is open in weak topology. The collection of sets $\{F_{\bfx}\ | \ \bfx\in X\}$ covers $X$. There exist a final cover of $X$ by sets $F_{\bfx}$:
\[
X\subset \bigcup_{j=1,\ldots, g} F_{\bfx_j}
\]
for some finite set $\{\bfx_1, \ldots, \bfx_l\} \subset X$. The composite predicate that separates $X$ from $Y$ can be chosen in the form $B(\bfz)=B_{\bfx_1}(\bfz)\lor \ldots \lor B_{\bfx_l}(\bfz)$. $\square$

\section{Discussion}\label{sec:discussion}

The nature of phenomenon described is universal and does not depend on many details of the distribution. In some sense, it should be just essentially high-dimensional. Due to  space limitations cannot review all these generalizations here but will nevertheless provide several examples.

\subsection{Equidistribution in an ellipsoid}\label{sec:discussion.ellipsoid}


Let points in the set $\mathcal{M}$ be selected by independent trials taken from the equidistribution in a $n$-dimensional ellipsoid. Without loss of generality, we present this ellipsoid in the orthonormal eigenbasis:

\begin{equation}\label{ellipsoid}
E_n=\{\bfx\in\Real^n | \sum_{i=1}^n \frac{x_i^2}{c_i^2}\leq 1\},
\end{equation}
where $r c_i$ are the semi-principal axes. The linear transformation
\[
(x_1,\ldots, x_n)\mapsto \left(\frac{x_1}{c_1},\ldots, \frac{x_n}{c_n}\right)
\]
transforms the ellipsoid into the unit ball. The volume of every set in the new coordinates scales with
the factor $1/\prod_{i}c_i$. Therefore the ratio of two volumes does not change, and the equidistribution in the ellipsoid is transformed into the equidistribution in the unit ball. Hyperplanes are transformed into hyperplanes and the property of linear separability is not affected by a nonsingular linear transformation. Thus, the following corollaries from Theorems \ref{theorem:point_separation}, \ref{theorem:all_points_separation}  hold.

\begin{cor}\label{cor:point_separation} Let $\mathcal{M}$ be formed by a finite number of i.i.d. trials taken from the equidistribution in a $n$-dimensional ellipsoid $E_n$ (\ref{ellipsoid}), and let $\bfx$ be a test point drawn independently from the same distribution. Then $\bfx$ can be separated from $\mathcal{M}$ by a linear functional with probability $P_1(\mathcal{M},n)$:
\[
P_1(\mathcal{M},n)\geq (1-(1-\varepsilon)^n)\left(1-\frac{\rho(\varepsilon)^n}{2}\right)^M.
\]
\end{cor}

\begin{cor}\label{cor:extremality} Let $\mathcal{M}$ be formed by a finite number of i.i.d. trials taken from the equidistribution in a $n$-dimensional ellipsoid $E_n$. With probability
\[
P_M(\mathcal{M},n)\geq  \left[(1-(1-\varepsilon)^n) \left(1-(M-1)\frac{\rho(\varepsilon)^n}{2}\right)\right]^M
\]
each point $\bfy$ from  the random set $\mathcal{M}$ can be separated from $\mathcal{M}\backslash \{\bfy\}$ by a linear functional.
\end{cor}

\subsection{Equidistributions in a cube and normal distributions}\label{sec:discussion.other_distributions}

It is well known that, for $n$ sufficiently large, samples drawn from an $n$-dimensional normal distribution  \cite{Levi1951} concentrate near a corresponding $n$-sphere. Similarly, samples generated from an equidistribution in an $n$-dimensional cube concentrate near a corresponding $n$-sphere too. In this respect, one might expect that the estimates derived in Theorems \ref{theorem:point_separation}-\ref{theorem:two_neuron_separation} hold for these distributions too, asymptotically in $n$. Our numerical experiments below illustrate that this is indeed the case.

\begin{figure*}[!th]
\begin{center}
\begin{minipage}[h]{\linewidth}
\centering
\small
\begin{tabular*}{\textwidth}{
@{\extracolsep{\fill}} |c|c|c|c|c|c|}
\hline
Dimension, $n$ & $2$ & $5$ & $10$ & $20$ & $30$\\
\hline
{\it Ball} &  &  & & & \\
{(theoretical estimate)} & $3.7 \cdot 10^{-5}$ & $0.0364$ & $0.4096$ & $0.9455$ & $ 0.9975$\\
 & & & & & \\
\hline
{\it Cube} &   & & & &  \\
min:    & $4\cdot 10^{-4}$ & $0.0089$  & $0.2580$ & $0.9408$ & $0.9986$ \\
median: & $4\cdot 10^{-4}$ & $0.0110$  & $0.2737$ & $0.9469$ & $0.9992$ \\
max:    & $5\cdot 10^{-4}$ & $0.0137$  & $0.2847$ & $0.9511$ & $1.0$\\
 & & & & & \\
\hline
 {\raggedleft{\it Gaussian}}  &  &  & & & \\
min:    & $5\cdot 10^{-4}$  & $0.0122$ & $0.1403$ & $0.7559$ & $0.9792$ \\
median: & $10\cdot 10^{-4}$ & $0.0153$ & $0.1568$ & $0.7698$ & $0.9817$ \\
max:    & $0.0014$          & $0.0183$ & $0.1778$ & $0.7836$ & $0.9848$ \\
 & & & & & \\
\hline
\end{tabular*}
\end{minipage}
\vspace{5mm}
\begin{minipage}[h]{0.5\linewidth}
\centering
\vspace{2mm}
\includegraphics[width=\linewidth]{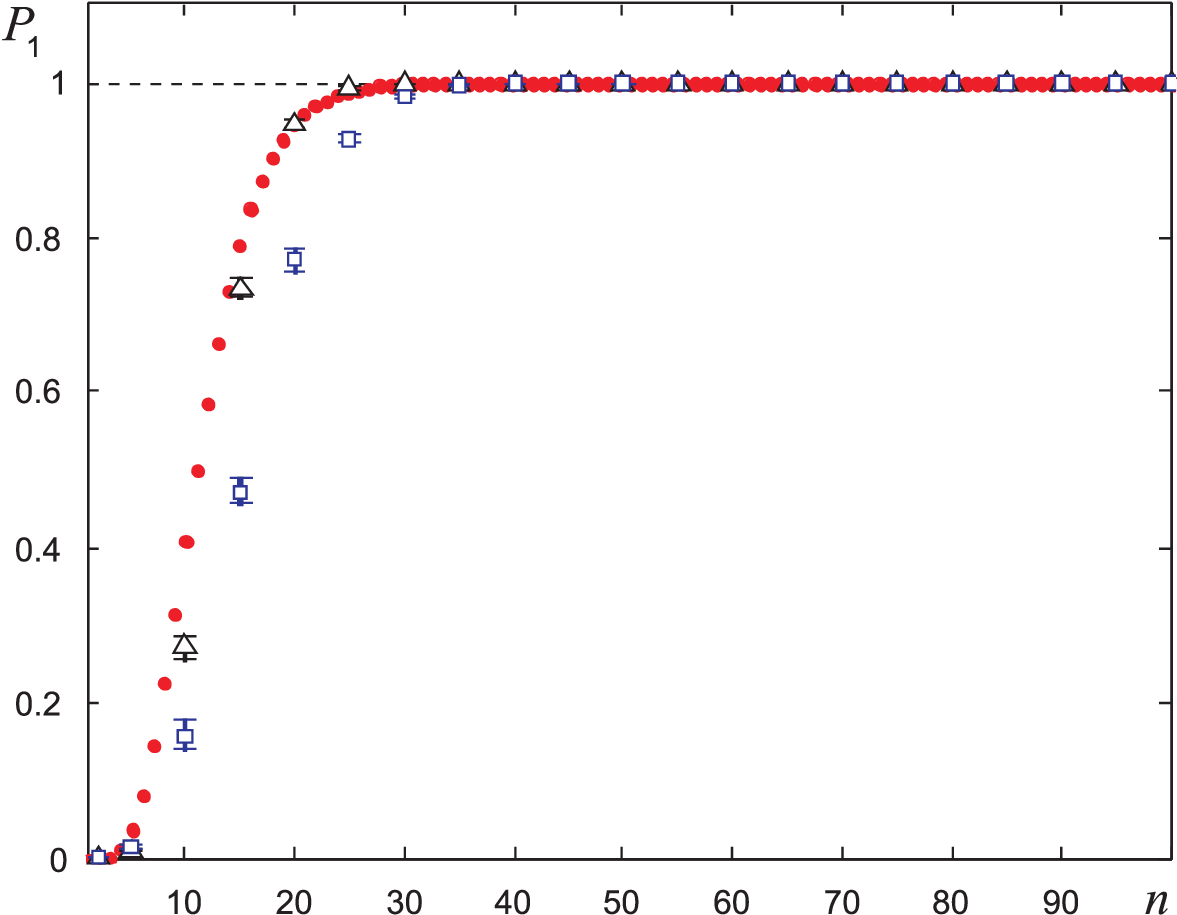}
\end{minipage}
\end{center}
\caption{Numerical and theoretical estimates of $P_1 (\mathcal{M},n)$ for various distributions and $n$. {\it Top panel}, top row: theoretical estimate of $P_1(\mathcal{M},n)$ for equidistributions in $n$-balls  derived in accordance with (\ref{eq:random_point_separation}) in the statement of Theorem \ref{theorem:point_separation}. {\it Left panel}, rows $2,3$: numerical estimates $F_1(\mathcal{M},n)$ of $P_1(\mathcal{M},n)$ for both normal and equidistribution in an $n$-cube for various values of $n$. {\it Bottom panel:} solid circles show the values of theoretical estimates $P_1(\mathcal{M},n)$, triangles show empirical means of $F_1(\mathcal{M},n)$ for the samples drawn from equidistribution in the cube $[-1,1]^{n}$, and squares correspond to empirical  means of $F_1(\mathcal{M},n)$ for the samples drown from the Gaussian (normal) distribution. Whiskers in the plots indicate maximal and minimal values in of $F_1(\mathcal{M},n)$ in each group of experiments.}\label{fig:separation_benchmarks}
\end{figure*}

The experiments were as follows. For each distribution (normal distribution in $\Real^n$ and equidistribution in the $n$-cube $[-1,1]^n$) and a given $n$ an i.i.d. sample $\mathcal{M}$ of $M=10^4$ vectors was drawn. For each vector $\bfy$ in this sample we constructed the functional  $l_{\bfy}(\bfx)= \langle \bfy,\bfx\rangle - \|\bfy\|^2$ (cf.  the proof of Lemma \ref{lem:given_point_separation}). For each $\bfy\in\mathcal{M}$ and  $\bfx\in\mathcal{M}$, $\bfx\neq \bfy$ the sign of $l_{\bfy}(\bfx)$ was evaluated, and the total number $N$ of instances when $l_{\bfy}(\bfx)< 0$ for all $\bfx\in\mathcal{M}$, $\bfx\neq \bfy$ was calculated. The latter number is a lower bound estimate of the number of points in $\mathcal{M}$ that are linearly separable from the rest in the sample. This was followed by deriving the values of the success frequencies, $F_1(\mathcal{M},n)=N/(M-1)$. For each $n$ the experiment was repeated $50$ times. Outcomes of these experiments are presented in Fig. \ref{fig:separation_benchmarks}
As we can see from Fig. \ref{fig:separation_benchmarks}, despite that the samples are drawn from different distributions, for $n>30$ these differences do not significantly affect point separability properties. This is due to pronounced influence of measure concentration in these dimensions.

\subsection{One trial non-iterative learning}

	Our basic model of linear {  separation} of a given query point $\bfy\in\mathcal{M}$ from any other $\bfx\in\mathcal{M}$, $\bfx\neq \bfy$ was
\begin{equation}\label{eq:spherical_cap_model}
l_{\bfy}(\bfx)=\left\langle \frac{\bfy}{\|\bfy\|}, \bfx\right\rangle-\|\bfy\|<0 \mbox{ for } \bfx\in \mathcal{M} \smallsetminus \{\bfy\}
\end{equation}
Deriving these functionals is a genuine one-shot procedure and does not require iterative learning. The construction, however, assumes that the distribution from which the sample $\mathcal{M}$ is drawn is close in some sense to an equidistribution in the unit ball $B_n(1)$.

In more general cases, e.g. when sampling from the equidistribution in an ellipsoid, the functionals $l_{\bfy}(\bfx)$ can be replaced with Fisher linear discriminants:
\begin{equation}\label{eq:Fisher_cap}
l_{\bfy}(\bfx)=\left\langle \frac{\bfw(\bfy,\mathcal{M})}{\|\bfw(\bfy,\mathcal{M})\|}, \bfx\right\rangle-c,
\end{equation}
where
\[
\bfw(\bfy,\mathcal{M})=\Cov(\mathcal{M})^{-1}\left(\bfy-{ \frac{1}{M-1}}\sum_{\bfx\neq \bfy} \bfx\right),
\]
{and $\Cov(\mathcal{M})$} is the non-singular covariance matrix of the sample $\mathcal{M}$, and $c$ is a parameter. The value of $c$ could be chosen as $c=\left\langle \frac{\bfw(\bfy,\mathcal{M})}{\|\bfw(\bfy,\mathcal{M})\|},\bfy\right\rangle$. The procedure for generating separating functionals $l_{\bfy}(\bfx)$ remains non-iterative, but it does require knowledge of the covariance matrix $\Sigma$.

Note that if $\mathcal{M}$ is centered at $\boldsymbol{0}$ then ${ \frac{1}{M-1}}\sum_{\bfx\neq \bfy} \bfx$ { is approximately $\boldsymbol{0}$}, and (\ref{eq:Fisher_cap}) reduces to (\ref{eq:spherical_cap_model}) after the corresponding Mahalanobis transformation: $\bfx\mapsto {\Cov(\mathcal{M})}^{-1/2} \bfx$. {  If the transformation $\bfx\mapsto {\Cov(\mathcal{M})}^{-1/2} \bfx$ transforms the ellipsoid from which the sample $\mathcal{M}$ is drawn into the unit ball then (\ref{eq:Fisher_cap}) becomes equivalent to (\ref{eq:spherical_cap_model}), and separation properties of Fisher discriminants (\ref{eq:Fisher_cap}) follow in the same way as stated in Corollaries \ref{cor:point_separation} and \ref{cor:extremality}.}

In addition, or as an alternative, whitening or decorrelation transformations could be applied to $\mathcal{M}$ too. In case the covariance matrix {$\Cov(\mathcal{M})$} is singular or ill-conditioned, projecting the sample $\mathcal{M}$ onto relevant Principal Components may be required.

\subsection{Extreme selectivity of biological neurons}

High separation capabilities of linear  functionals (Theorems \ref{theorem:point_separation}--\ref{theorem:two_neuron_separation}, Remark \ref{rem:ExpSample}) may shed light on the puzzle associated with  the so-called Grandmother \cite{Quiroga:2005}, \cite{Quiroga:2009} or Concept cells \cite{Quiroga:2012} in neuroscience.  In the center of this puzzle is the wealth of empirical evidence that  some neurons in human brain respond unexpectedly selective to particular persons or objects. Not only brain is able to respond selectively to “rare” individual stimuli but also such selectivity can be acquired very rapidly from limited number of experiences \cite{Quiroga:2015}. The question is: Why small ensembles of neurons may deliver such a sophisticated functionality reliably? As follows from Theorems \ref{theorem:point_separation}--\ref{theorem:two_neuron_separation}, observed extreme selectivity of neuronal responses may be attributed to the very nature of basic physiological mechanisms in neurons involving calculation of linear functionals (\ref{eq:linear_func}).

\section{Examples}\label{sec:examples}

Let us now illustrate our theoretical results with examples of correcting legacy AI systems by linear functionals (\ref{eq:linear_func}) and their combinations. As a legacy AI system we selected a Convolutional Neural Network trained to detect objects in images. Our choice of the legacy system was motivated by that these networks demonstrate remarkably strong performance on benchmarks and, reportedly, may outperform humans on popular classification tasks such as ILSVRC \cite{He:2016}. Despite this, as has been mentioned earlier, even for the most current state of the art CNNs false positives are not uncommon. Therefore, we investigate in the following experiments if it is possible to take one of these cutting edge networks and train a {\it one-neuron} filter to eliminate all the false positives produced. We will also look at what effect, if any, this filter will have on true positive numbers.

\subsection{Training the CNN}

For these experiments we trained the VGG-11 convolutional network to be our classifier \cite{Simonyan:2015}. Instead of the 1000 classes of Imagenet we trained it to perform the simpler task of identifying just one object class: pedestrians.

The VGG network was chosen for both its simple homogeneous architecture as well as its proven classification ability at recent Imagenet competitions \cite{Russ:2015}. We chose to train the VGG-11 $A$ over the deeper $16$ and $19$ layer VGG networks due to hardware and time constraints.

\textit{Datasets}: In order to train the network a set of $114,000$ positive pedestrian and $375,000$ negative non-pedestrian RGB images, re-sized to $128\times128$, were collected and used as a {\it training set}. Our {\it testing set} comprised of further $10, 000$ positives and $10,000$ negatives. The training and testing sets had no common elements.

The training procedure and choice of hyper parameters for the network follows largely from \cite{Simonyan:2015}. but with some small adjustments to take into account the reduced number of classes being detected. We set momentum to $0.9$ and used a mini batch size of $32$.

Our initial learning rate was set to $0.00125$ and this rate was reduced by a factor of $10$ after $25$ epochs and again after $50$ epochs. Dropout regularisation was used for the first two fully connected layers with a ratio of $0.5$. To initialise weights we used the Xavier initialisation \cite{Glorot:2010} which we found helped training to converge faster. {After $75$ epochs, the validation accuracy started to fall and, we stopped the training at $75$ epochs} in order to avoid overfitting.

Training the network was done using the publicly available Caffe deep learning framework \cite{Caffe}. For the purposes of evaluating performance of the network, our null-hypothesis was that {\it no pedestrians are present in the scanning window}. True positive hence is an image crop (formed by the scanning window) containing a pedestrian and for which the CNN reports such a presence, false positive is an image crop for which the CNN reports presence of a pedestrian whilst none are present. True negatives and false negatives are defined accordingly. On the training set, the VGG-11 convolutional network showed $100\%$ correct classification performance:  the rates of true positives and true negatives were all equal to one.

\subsection{Training {  one-neuron} legacy AI system correctors}

\subsubsection{Error detection and selection of features}

First, a multi-scale sliding window approach is used on each video frame to provide proposals that can be run through the trained CNN. These proposals are re-sized to $128\times128$ and passed through the CNN for classification, non-maximum suppression is then applied to the positive proposals. Bounding boxes are drawn and we compared results to a ground truth so we can identify false positives.

For each positive and false positive (and its respective set of proposals before non-maximum suppression) we extracted the second to last fully connected layer from the CNN. These extracted feature vectors have dimension $4096$.   We used these feature vectors to construct {\it one-neuron} correctors  filtering out the false positives that have been manually identified. These we will refer to in the text as 'trash models'.

\subsubsection{Spherical cap model, (\ref{eq:spherical_cap_model})}

The first and the most simple trash model filter that we constructed was the spherical cap model as specified by (\ref{eq:spherical_cap_model}). In this model, after centering the training data, the query points $y$, adjusted for the mean of the training data, were the actual false positives detected.

\subsubsection{Fisher cap model, (\ref{eq:Fisher_cap})} 
\begin{figure}
\centering
\includegraphics[width=0.6\linewidth]{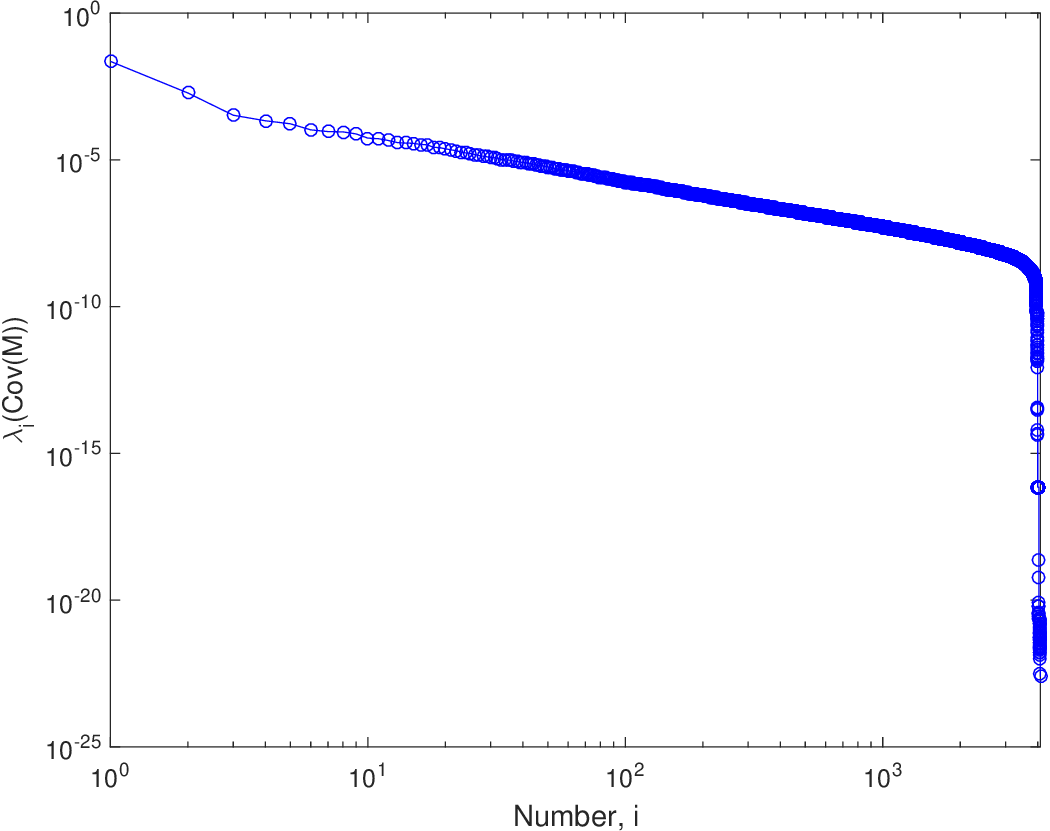}
\caption{Log-log plot of the eigenvalues $\lambda_i({\Cov(\mathcal{M}))}$ of the training data covariance matrix {$\Cov(\mathcal{M})$}.}\label{fig:eigenvalues}
\end{figure}

Our second model type was the Fisher cap model (\ref{eq:Fisher_cap}). The original model involves the covariance matrix {$\Cov(\mathcal{M})$} corresponding to $4096$ dimensional feature vectors of positives in the training set. Examination of this matrix revealed that the matrix is extremely ill-conditioned and hence (\ref{eq:Fisher_cap}) cannot be used unless the problem is regularized. To overcome the singularity issue the PCA-based dimensionality reduction was employed. To estimate the number of relevant principal components needed we performed the broken stick  \cite{McArthur:1957} and the Kaiser-Guttman \cite{Jackson:1993} tests. The broken stick criterion returned the value of $13$, and the Kaiser rule returned $45$. The distribution of the eigenvalues of {$\Cov(\mathcal{M})$} is shown in Fig. \ref{fig:eigenvalues}. We used these data to select lower and upper bound estimates of the number of feasible principal components needed. For the purposes of our experiments these were chosen to be $50$ and $2000$, respectively.


\subsubsection{Support Vector Machine (SVM)}


As the ultimate linear separability benchmark, classical linear SVM model has been used. It was trained using {the standard} liblinear software package \cite{Fan:2008}\footnote{ {The package is available at GitHub \url{https://github.com/cjlin1/liblinear}.}} on our two sets of normalised CNN feature vectors. {No kernel transformations have been used in the process. When training the SVM models we used  $L_2$-regularized $L_2$-loss function for the dual (option `-s 1'), set termination tolerance to $0.001$ (option `-e 0.001') and  bias parameter value to $1$ (option `-B 1'). Classes have been weighted and treated equally (options `-w1 1' and `-w-1 1'), and the value of the regularization parameter, parameter $C$ in \cite{Fan:2008}, was varying from $1$ to $110$ depending on the outcomes of classification (the value was set through the option `-c '). We} found that as the number of false positives being trained on increased it was necessary to also increase the value of the $C$ parameter to maintain a perfect separation of the training points.

\subsection{Implementation}

At test time trained linear corrector models (trash model filters) were placed at the end of our detection pipeline, each one at a time. For any proposal given a positive score by our CNN we again extract the second to last fully connected layer.  The CNN feature vectors from these positive proposals are then run through the trash model filter.

Any detection that then gives a positive score from the linear trash model is consequently removed by turning its detection score negative.

\subsection{Results}

\textit{Test videos}: For our experiments we used the original training and testing sets as well as three different videos (not used in the training set generation) to test trash model creation and its effectiveness at removing false positives.  The first sequence is the NOTTINGHAM video \cite{Nottingham} that we created ourselves from the streets of Nottingham consisting of $435$ frames taken with an action camera. The second video is the INRIA test set consisting of $288$ frames \cite{Dalal:2005} , the third is the LINTHESCHER sequence produced by ETHZ consiting of $1208$ frames \cite{Ess:2008}.
For each test video a new trash model was trained.

\subsubsection{Spherical cap model, (\ref{eq:spherical_cap_model})}

To test performance of this na\"ive model we run the trained CNN classifier through the testing set and identified false positives. From the set of false positives, candidates were selected at random and correcting trash model filters (\ref{eq:spherical_cap_model}) constructed.  Performance of the resulting systems was then evaluated on the sets comprising of $114,000$ positives (from the training set) and $1$ negative (the false positive identified). These tests returned true negative rates equal to $1$, as expected, and $0.978884$ true positive rate averaged over $25$ experiments. This is not surprising given that the model does not account for any statistical properties of the data apart from the assumption that the sample is taken from an equidistribution in $B_n(1)$. The latter assumption, as is particularly evident from Fig. \ref{fig:eigenvalues}, does not hold either. {  Equidistribution in an ellipsoid (or the multidimensional normal distribution, which is essentially the same) is a much better model of the data, and hence one would expect that Fisher discriminants (\ref{eq:Fisher_cap}) would perform better. We shall see now that this is indeed the case.}

\subsubsection{Fisher cap model, (\ref{eq:Fisher_cap})}

The Fisher discriminant is {  essentially} the functional we use in theoretical estimate for distributions in ellipsoid. It is the same spherical cup, {  albeit after the whitening applied}. The Fisher cap models were first constructed and tested on the data used to train and test the original CNN classifier. We started with the data projected onto the first $50$ principal components. The $25$ false positive candidates, taken from the testing set, were chosen at random as described in the previous section. {  They modeled {\it single}} mistakes of the legacy classifier. The models were then prepared using the covariance matrices of the original training data set. On the data sets comprising of $114,000$ positives from the training set and $1$ negative corresponding to the single randomly chosen identified false positive all models delivered true negative and true positive rates equal to $1$. The result persisted when the number of principal components used was increased to $2000$.

In the next set of experiments we {investigated numerically} if a single neuron with the weights given by the Fisher discriminant formula would be capable of filtering out more than just one false positive candidate. {We note that the question is not trivial even if the two sets are linearly separable: separability by Fisher linear discriminants imply linear separability, but the converse does not necessarily holds true.} For this purpose we randomly selected {  several} samples $\mathcal{Y}$ of $25$ false positive candidates from the training data set and constructed models (\ref{eq:Fisher_cap}) aiming at separating all $25$ false positives in each sample from $114,000$ true positives in the data set. The weights $\bfw$ of model (\ref{eq:Fisher_cap}) depended on the sets of false positives $\mathcal{FP}$ and the set $\mathcal{TP}$ of true positives as follows:
\[
\begin{split}
\bfw(\mathcal{FP},\mathcal{TP})=&\left({\Cov(\mathcal{TP})} + {\Cov(\mathcal{FP})}\right)^{-1} \left(\frac{1}{|\mathcal{FP}|}\sum_{\bfy\in\mathcal{FP}} \bfy -\frac{1}{|\mathcal{TP}|}\sum_{\bfx\in \mathcal{TP}} \bfx\right)
\end{split}
\]
where ${\Cov(\mathcal{TP})}$, ${\Cov(\mathcal{FP})}$ are the {empirical} covariance matrices of true and false positives, respectively, and the value of parameter $c$ was chosen as
\[
c=\min_{\bfy\in\mathcal{FP}} \left\langle \frac{\bfw(\mathcal{FP},\mathcal{TP})}{\|\bfw(\mathcal{FP},\mathcal{TP})\|},\bfy \right\rangle.
\]
We did not observe perfect separation if only the first $50$ principal components were used to construct the models. Perfect separation, however, was persistently observed when the number of first principle components being used exceeded $87$. {  For data projected on more than the first $87$ principal components one neuron with weights selected by the Fisher linear discriminant formula corrected $25$ errors without doing any damage to classification capabilities (original skills) of the legacy AI system on the training set.}

{The difference in performance with dimension can be explained by that the hyperplane corresponding to Fisher linear discriminants for the $25$ randomly chosen false positives appears to be closer to the origin than each individual data point in this set.  In terms of the theoretical argument presented in Theorems \ref{theorem:point_separation}, \ref{theorem:all_points_separation} this can be seen as an increase in the value of $\varepsilon$ corresponding to the test point. This increase leads to the increase of the value of $\rho(\varepsilon)^{n}/2$ in (\ref{eq:single_neuron_separation_at_epsilon}) and results in reduced values of the probability of separation. When the dimensionality of the data, $n$, grows, the term $\rho(\varepsilon)^{n}/2$ decreases mitigating the effect. This is consistent with what we observed in this experiment. We expect that the observed negative consequences affecting linear separability of multiple points could be addressed via employing a two-neuron separation as well.}

\begin{figure*}[ht]
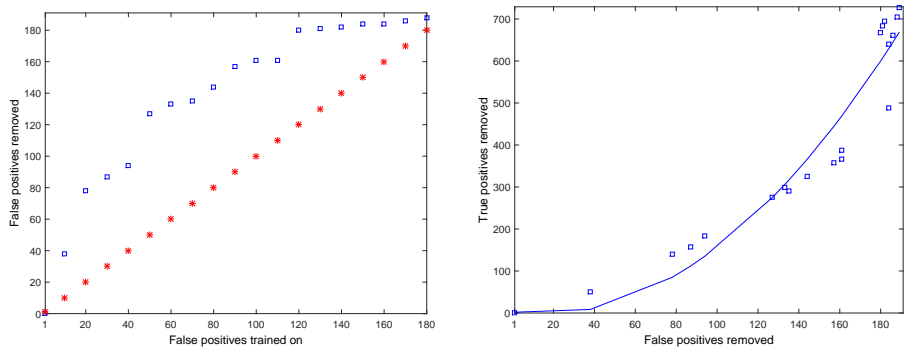

\centering
\includegraphics[width=0.47\textwidth]{nottingham_results_fisher_fp_vs_trained.eps} \hspace{3mm} \includegraphics[width=0.47\textwidth]{nottingham_results_fisher_tp_vs_fp.eps}
\vspace{1mm}
\caption{Performance of {  {\it one-neuron} corrector built using} Fisher cap model (\ref{eq:Fisher_cap}). {\it Left panel}: the number of false positives removed  as a function of the number of false positives the model was built on. Stars indicate the number of false positives used for building the model. Squares correspond to the number of false positives removed by the model. {\it Right panel}: the number of true positives removed as a function of the number of false positives removed. The actual measurements are shown as squares. Solid line is the least-square fit of a quadratic.}\label{fig:fp_fisher_effect}
\vspace{5mm}
\end{figure*}

In the third group of tests we built single neuron trash models (\ref{eq:Fisher_cap}) on varying numbers of false positives from the NOTTINGHAM video. The true positives (totalling to $2896$) and  the false positives ($189$) in this video differed from those in the training set. The original training data set was projected onto the first $2000$ principal components. Typical performance of the legacy CNN system with the corrector is shown in Fig. \ref{fig:fp_fisher_effect}.
As we can see from this figure, single false positives are dealt with successfully without any detrimental affects on the true positive rates. Increasing the number of false positives that the single neuron trash model is to cope with resulted in gradual deterioration of the true positive rates. One may expect that, according to Theorem \ref{theorem:two_neuron_separation}, increasing of the number of neurons in trash models could significantly improve the power of correctors. Conducting these experiments, however, was outside of the main focus of this example.

\subsubsection{SVM model}

The Spherical cap and the Fisher cap models are by no means the ultimate tests of separability. Their main advantage is that they are purely non-iterative. To test extremal performance of linear functionals more sophisticated methods for their construction, such as e.g. SVMs, are needed. To assess this capability we trained SVM trash models on varying numbers of false positives from the NOTTINGHAM video, as a benchmark. Results are plotted in Fig. \ref{fig:fp_svm_effect} below. The true positives were taken from the CNN training data set.
\begin{figure*}
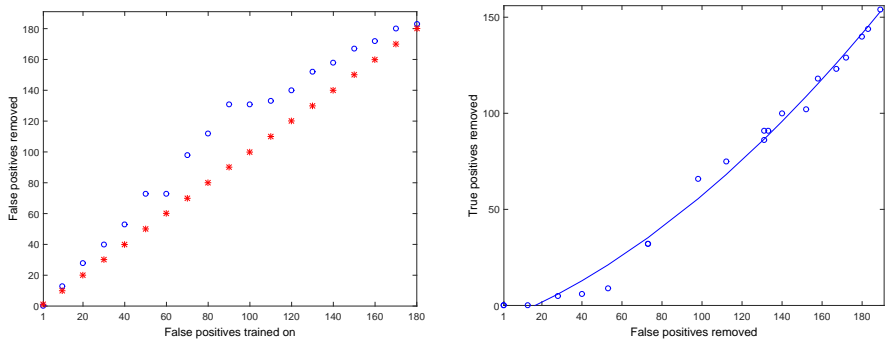

\centering
\includegraphics[width=0.46\textwidth]{nottingham_results_svm_fp_vs_trained.eps} \hspace{3mm} \includegraphics[width=0.46\textwidth]{nottingham_results_svm_tp_vs_fp.eps}
\vspace{2mm}
\caption{Performance of { {\it one-neuron} corrector constructed using} the linear SVM model. {\it Left panel}: the number of false positives removed as a function of the number of false positives the model was trained on. Stars indicate the number of false positives used for training the model. Circles correspond to the number of false positives removed by the model {\it Right panel}: the number of true positives removed as a function of the number of false positives removed. The actual measurements are presented as circles. Solid line is the least-square fit of a quadratic.}\label{fig:fp_svm_effect}
\end{figure*}
The SVM trash model successfully removes $13$ false positives without affecting the true positives rate. Its performance deteriorates at a much slower rate than for the Fisher cap model. This, however, is balanced by the fact that the Fisher cap model removed significantly more false positives than it was trained on.

Similar level of performance was observed for other testing videos, including INRIA and ETH LINTHESCHER videos. The results are provided in Tabes \ref{table:INRIA} and \ref{table:ETH} below.
\begin{table}[h!]
\caption{CNN performance on the INRIA video with and without trash model filtering}\label{table:INRIA}
\begin{center}
\begin{tabular}{ | m{6em} | m{2.5cm}| m{2.5cm} | }
\hline
& Without trash model & With SVM trash model \\
\hline
True positives & 490 & 489 \\
\hline
False positives & 31 & 0 \\
\hline
\end{tabular}
\end{center}
\end{table}

\begin{table}[h!]
\caption{CNN performance on the ETH LINTHESCHER video with and without trash model filtering}\label{table:ETH}
\begin{center}
\begin{tabular}{ | m{6em} | m{2.5cm}| m{2.5cm} | }
\hline
& Without trash model & With SVM trash model \\
\hline
True positives & 4288 & 4170 \\
\hline
False positives & 9 & 0 \\
\hline
\end{tabular}
\end{center}
\end{table}
%

\subsection{Results summary}

From the results in Figures \ref{fig:fp_fisher_effect}, \ref{fig:fp_svm_effect} and Tables \ref{table:INRIA}, \ref{table:ETH}  we have demonstrated that it is possible to train a simple single-neuron corrector capable of rectifying mistakes of legacy AI systems. As can be seen from Figures \ref{fig:fp_fisher_effect}, \ref{fig:fp_svm_effect} this filtering of false positives comes at a varying cost to the true positive detections. As we increase the number of false positives that we train the trash model on we see the number of true positives removed starts to rise as well. Our tests illustrate, albeit empirically, that the separation theorems stated in this paper are viable in applications. What is particularly remarkable is that we were able to remove more than $10$ false positives at no cost to true positive detections in the NOTTINGHAM video by the use of a single linear function.

{  From practical perspectives, the main operational domain of the correctors is likely to be near the origin along the `removed false positives' axis. In this domain generalization capacities of Fisher discriminant are significantly higher than those of the SVMs. As can be seen from Fig. \ref{fig:fp_fisher_effect}, Fig. \ref{fig:fp_svm_effect}, left panels, Fisher discriminants are also efficient in 'cutting off' significantly more mistakes than they have been trained on. In this operational domain near the origin, they do so without doing much damage to the legacy AI system. SVMs perform better for larger values of false positives. This is not surprising as their training is iterative and accounts for much more information about the data than simply sample covariance and mean.}

\section{Conclusion}\label{sec:conclusion}

Stochastic separation theorems allowed us to create non-iterative one-trial correctors for legacy AI systems. These correctors were tested on benchmarks and on real-life industrial data and  outperformed the theoretical estimates. In these experiments, already one-neuron correctors fixed more than ten independent mistakes  without   doing any damage  to existing skills of the legacy AI system. Not only these experiments illustrated application of the approach, but also they revealed importance of whitening and regularization for successful correction. In order to reduce the impact on true positive detections a two neuron cascade classifier could be trained as a corrector. Further elaboration of the theory of  non-iterative one-trial correctors for legacy AI systems and testing them in applications will be the subject of our future work.

The stochastic separation theorems are proven for large sets of randomly drawn i.i.d. data in high dimensional space. Each point can be separated from the rest of the sample   by a hyperplane, with high probability. In a convex hull of a random finite set, all points of the set are vertexes (with high probability). These results are valid even for exponentially large samples. The estimates of the separation probability and allowable sample size age given by Eqs.  (\ref{eq:random_point_separation}), (\ref{eq:capacity}),  (\ref{eq:all_random_points_separation}), and (\ref{eq:capacityExtrem}). It is not much surprising that the separation by small ensembles of uncorrelated neurons is much more efficient than the separation by single neurons (Figure~\ref{fig:two_neuron_separation_max}).

We showed that high dimensionality of data can play a major and positive role in various machine learning and data analysis tasks, including problems of separation, filtration, and selection.

	In contrast to naive intuition suggesting that high dimensionality of data more often than not brings in additional complexity and uncertainty, our findings contribute to the idea that high data dimensionality may constitute a valuable blessing too. The term `blessing of dimensionality' was used in \cite{Anderson2014} to describe separation by Gaussians mixture models or spheres, which are equivalent in high dimension. This blessing, formulated here in the form of several stochastic separation theorems, offers several new insights for big data analysis. This is a part of measure concentration phenomena \cite{Gromov:1999,GAFA:Gromov:2003,milman2009asymptotic} which form the background of classical statistical physics (Gibbs theorem about equivalence of microcanonic and canonic ensembles \cite{Gibbs1902}) and asymptotic theorems in probability \cite{Levi1951}. In machine learning, these phenomena are in the background of the random projection methods \cite{johnson1984extensions}, learning large Gaussian mixtures \cite{Anderson2014}, and various randomized approaches to learning \cite{Wang:2016}, \cite{scardapane2017randomness} and bring light in the theory of approximation by random bases \cite{GorTyu:2016}. It is highly probable that  the recently described manifold disentanglement effects \cite{Brahma2016} are universal in essentially high dimensional data analysis and relate to essential multidimensionality  rather than to specific deep learning algorithms.    This may be the manifestation of concentration near the parts of the boundary with the lowest curvature.

The plan of further technical development of the theory and methods presented in the paper is clear:
\begin{itemize}
\item To prove the separation theorems in much higher generality: for sufficiently regular essentially multidimensional probability distribution and a random set of  weakly dependent (or independent)  samples every point is linearly separated from the rest of the set  with high probability.
\item To find explicit estimation of the allowed sample size in general case.
\item To elaborate the theorems and estimates in high generality for separation by small neural networks.
\item To develop and test generators of correctors for various standard AI tasks.
\end{itemize}

But the most inspiring consequence of the measure concentration phenomena is the paradigm shift. It is a common point of view that the complex learning systems should produce {\em complex} knowledge and skills. On contrary,  {  it} seems to be possible that the main function of many learning system, both technical and biological, in addition to production of {\em simple} skills, is a special preprocessing. They transform the input flux (`reality') into essentially multidimensional and quasi-random distribution of signals and images plus, may be, some simple low dimensional and more regular signal. After such a transformation, ensembles of non-interacting or weakly interacting small neural networks (`correctors' of simple skills) can solve complicated problems.

\section*{References}
\bibliography{high-dimensional_data_analysis_journal}

\begin{thebibliography}{51}
\providecommand{\natexlab}[1]{#1}
\providecommand{\url}[1]{\texttt{#1}}
\expandafter\ifx\csname urlstyle\endcsname\relax
  \providecommand{\doi}[1]{doi: #1}\else
  \providecommand{\doi}{doi: \begingroup \urlstyle{rm}\Url}\fi

\bibitem[Anderson et~al.(2014)Anderson, Belkin, Goyal, Rademacher, and
  Voss]{Anderson2014}
J.~Anderson, M.~Belkin, N.~Goyal, L.~Rademacher, and J.~Voss.
\newblock The more, the merrier: the blessing of dimensionality for learning
  large {G}aussian mixtures.
\newblock \emph{Journal of Machine Learning Research: Workshop and Conference
  Proceedings}, 35:\penalty0 1--30, 2014.

\bibitem[Ball(1997)]{ball1997elementary}
K.~Ball.
\newblock An elementary introduction to modern convex geometry.
\newblock \emph{Flavors of geometry}, 31:\penalty0 1--58, 1997.

\bibitem[Bennett(1995)]{Legacy:1995}
K.~Bennett.
\newblock Legacy systems: coping with success.
\newblock \emph{IEEE Software}, 12\penalty0 (1):\penalty0 19--23, 1995.

\bibitem[Bisbal et~al.(1999)Bisbal, Lawless, Wu, and Grimson]{Legacy:1999}
J.~Bisbal, D.~Lawless, B.~Wu, and J.~Grimson.
\newblock Legacy information systems: issues and directions.
\newblock \emph{IEEE Software}, 16\penalty0 (5):\penalty0 103--111, 1999.

\bibitem[Brahma et~al.(2016)Brahma, Wu, and She]{Brahma2016}
P.~P. Brahma, D.~Wu, and Y.~She.
\newblock Why deep learning works: a manifold disentanglement perspective.
\newblock \emph{IEEE Transactions On Neural Networks And Learning Systems},
  27\penalty0 (10):\penalty0 1997--2008, 2016.

\bibitem[Burton(2016)]{Nottingham}
R.~Burton.
\newblock Nottingham video, 2016.
\newblock URL \url{https://youtu.be/SJbhOJQCSuQ}.
\newblock A test video for pedestrians detection taken from the streets of
  Nottingham by an action camera.

\bibitem[Chen et~al.(2015)Chen, Li, Li, Lin, Wang, Wang, Xiao, Xu, Zhang, and
  Zhang]{MXNet}
T.~Chen, M.~Li, Y.~Li, M.~Lin, N.~Wang, M.~Wang, T.~Xiao, B.~Xu, C.~Zhang, and
  Z.~Zhang.
\newblock Mxnet: A flexible and efficient machine learning library for
  heterogeneous distributed systems.
\newblock https://github.com/dmlc/mxnet, 2015.

\bibitem[Dalal and Triggs(2005)]{Dalal:2005}
N.~Dalal and B.~Triggs.
\newblock Histograms of oriented gradients for human detection.
\newblock In \emph{Proc. of the IEEE Conference on Computer Vision and Pattern
  Recognition}, pages 886--893, 2005.

\bibitem[Ess et~al.(2008)Ess, Leibe, Schindler, and van Gool]{Ess:2008}
A.~Ess, B.~Leibe, K.~Schindler, and L.~van Gool.
\newblock A mobile vision system for robust multi-person tracking.
\newblock In \emph{Proc. of the IEEE Conference on Computer Vision and Pattern
  Recognition}, pages 1--8, 2008.
\newblock {DOI}: 10.1109/CVPR.2008.4587581.

\bibitem[Fan et~al.(2008)Fan, Chang, Hsieh, Wang, and Lin]{Fan:2008}
R.-E. Fan, K.-W. Chang, C.-J. Hsieh, X.-R. Wang, and C.-J. Lin.
\newblock Liblinear: A library for large linear classification.
\newblock \emph{Journal of Machine Learning Research}, 9:\penalty0 1871--1874,
  2008.
\newblock https://github.com/cjlin1/liblinear.

\bibitem[Fricker(2016)]{Science:2016:FP}
R.~J. Fricker.
\newblock False positives are statistically inevitable.
\newblock \emph{Science}, 351:\penalty0 569--570, 2016.

\bibitem[Gear and Kevrekidis(2005)]{Kevrekidis}
C.~Gear and I.~Kevrekidis.
\newblock Constraint-defined manifolds: a legacy code approach to
  low-dimensional computation.
\newblock \emph{Journal of Scientific Computing}, 25\penalty0 (1):\penalty0
  17--28, 2005.

\bibitem[Gibbs(1960 (1902))]{Gibbs1902}
J.~Gibbs.
\newblock \emph{Elementary Principles in Statistical Mechanics, developed with
  especial reference to the rational foundation of thermodynamics}.
\newblock Dover Publications, New York, 1960 (1902).

\bibitem[Glorot and Bengio(2010)]{Glorot:2010}
X.~Glorot and Y.~Bengio.
\newblock Understanding the difficulty of training deep feedforward neural
  networks.
\newblock In \emph{Proc. of the 13th International Conference on Arificial
  Intelligence and Statistics (AISTATS)}, volume~9, pages 249--256, 2010.

\bibitem[Gorban(2007)]{Gorban:2007}
A.~Gorban.
\newblock Order-disorder separation: Geometric revision.
\newblock \emph{Physica A}, 374:\penalty0 85--102, 2007.

\bibitem[Gorban et~al.(2016{\natexlab{a}})Gorban, Tyukin, Prokhorov, and
  Sofeikov]{GorTyu:2016}
A.~Gorban, I.~Tyukin, D.~Prokhorov, and K.~Sofeikov.
\newblock Approximation with random bases: Pro et contra.
\newblock \emph{Information Sciences}, 364--365:\penalty0 129--145,
  2016{\natexlab{a}}.

\bibitem[Gorban et~al.(2016{\natexlab{b}})Gorban, Tyukin, and
  Romanenko]{GorTyuRom2016}
A.~Gorban, I.~Tyukin, and I.~Romanenko.
\newblock The blessing of dimensionality: Separation theorems in the
  thermodynamic limit, 09 2016{\natexlab{b}}.
\newblock A talk given at TFMST 2016, 2nd IFAC Workshop on Thermodynamic
  Foundations of Mathematical Systems Theory. September 28-30, 2016, Vigo,
  Spain.

\bibitem[Gromov(1999)]{Gromov:1999}
M.~Gromov.
\newblock \emph{Metric Structures for Riemannian and non-Riemannian Spaces.
  With appendices by M. Katz, P. Pansu, S. Semmes. Translated from the French
  by Sean Muchael Bates}.
\newblock Birkhauser, Boston, MA, 1999.

\bibitem[Gromov(2003)]{GAFA:Gromov:2003}
M.~Gromov.
\newblock Isoperimetry of waists and concentration of maps.
\newblock \emph{GAFA, Geomteric and Functional Analysis}, 13:\penalty0
  178--215, 2003.

\bibitem[Halmos(1974)]{Halmos:1974}
P.~Halmos.
\newblock \emph{Finite dimensional vector spaces}.
\newblock Undergraduate Texts in Mathematics. Springer, 1974.

\bibitem[Hansen and Salamon(1990)]{Hansen:1990}
L.~K. Hansen and P.~Salamon.
\newblock Neural network ensembles.
\newblock \emph{{IEEE} {T}rans. on Patt. Analyis and Mach. Intel.}, 12\penalty0
  (10):\penalty0 993--1001, 1990.

\bibitem[He et~al.(2016)He, Zhang, Ren, and Sun]{He:2016}
K.~He, X.~Zhang, S.~Ren, and J.~Sun.
\newblock Deep residual learning for image recognition.
\newblock In \emph{Proc. of the IEEE Conference on Computer Vision and Pattern
  Recognition}, pages 770--778, 2016.
\newblock arXiv:1512.03385.

\bibitem[Ho(1995)]{Ho:1995}
T.~K. Ho.
\newblock Random decision forests.
\newblock In \emph{Proc. of the 3rd International Conference on Document
  Analysis and Recognition}, pages 993--1001, 1995.

\bibitem[Ho(1998)]{Ho:1998}
T.~K. Ho.
\newblock The random subspace method for constructing decision forests.
\newblock \emph{{IEEE} {T}rans. on Patt. Analyis and Mach. Intel.}, 20\penalty0
  (8):\penalty0 832--844, 1998.

\bibitem[Ison et~al.(2015)Ison, Quian~Quiroga, and Fried]{Quiroga:2015}
M.~Ison, R.~Quian~Quiroga, and I.~Fried.
\newblock Rapid encoding of new memories by individual neurons in the human
  brain.
\newblock \emph{Neuron}, 87\penalty0 (1):\penalty0 220--230, 2015.

\bibitem[Jackson(1993)]{Jackson:1993}
D.~Jackson.
\newblock Stopping rules in principal components analysis: A comparison of
  heuristical and statistical approaches.
\newblock \emph{Ecology}, 74\penalty0 (8):\penalty0 2204--2214, 1993.

\bibitem[Jia(2013)]{Caffe}
Y.~Jia.
\newblock Caffe: An open source convolutional architecture for fast feature
  embedding.
\newblock http://caffe.berkeleyvision.org/, 2013.

\bibitem[Johnson and Lindenstrauss(1984)]{johnson1984extensions}
W.~B. Johnson and J.~Lindenstrauss.
\newblock Extensions of lipschitz mappings into a hilbert space.
\newblock \emph{Contemporary mathematics}, 26\penalty0 (1):\penalty0 189--206,
  1984.

\bibitem[Krein and Milman(1940)]{Krein-Milman:1940}
M.~Krein and D.~Milman.
\newblock On extreme points of regular convex sets.
\newblock \emph{StudiaMath}, 9:\penalty0 133--138, 1940.

\bibitem[Krizhevsky et~al.(2012)Krizhevsky, Sutskever, and
  Hinton]{NIPS2012_4824}
A.~Krizhevsky, I.~Sutskever, and G.~Hinton.
\newblock Imagenet classification with deep convolutional neural networks.
\newblock In F.~Pereira, C.~J.~C. Burges, L.~Bottou, and K.~Q. Weinberger,
  editors, \emph{Advances in Neural Information Processing Systems 25}, pages
  1097--1105. Curran Associates, Inc., 2012.

\bibitem[Kuznetsova et~al.(2015)Kuznetsova, Hwang, Rosenhahn, and
  Sigal]{Kuznetsova:2015}
A.~Kuznetsova, S.~Hwang, B.~Rosenhahn, and L.~Sigal.
\newblock Expanding object detector’s horizon: Incremental learning framework
  for object detection in videos.
\newblock In \emph{Proc. of the IEEE Conference on Computer Vision and Pattern
  Recognition (CVPR)}, pages 28--36, 2015.

\bibitem[L\'evy(1951)]{Levi1951}
P.~L\'evy.
\newblock \emph{Probl\`emes concrets d'analyse fonctionnelle}.
\newblock Gauthier-Villars, Paris, second edition, 1951.

\bibitem[Macarthur(1957)]{McArthur:1957}
R.~Macarthur.
\newblock On the relative abundance of bird species.
\newblock \emph{Proc. Nat. Acad. of Sci.}, 43\penalty0 (3):\penalty0 293--295,
  1957.

\bibitem[Milman and Schechtman(2009)]{milman2009asymptotic}
V.~D. Milman and G.~Schechtman.
\newblock \emph{Asymptotic theory of finite dimensional normed spaces:
  Isoperimetric inequalities in riemannian manifolds}, volume 1200 of
  \emph{Lecture Notes in Mathematics}.
\newblock Springer, 2009.

\bibitem[Misra et~al.(2015)Misra, Shrivastava, and Hebert]{Misra:2015}
I.~Misra, A.~Shrivastava, and M.~Hebert.
\newblock Semi-supervised learning for object detectors from video.
\newblock In \emph{Proc. of the IEEE Conference on Computer Vision and Pattern
  Recognition (CVPR)}, pages 3594--3602, 2015.

\bibitem[Nguyen et~al.(2015)Nguyen, Yosinski, and Clune]{Nguyen:2015}
A.~Nguyen, J.~Yosinski, and J.~Clune.
\newblock Deep neural networks are easily fooled: High confidence predictions
  for unrecognizable images.
\newblock In \emph{Proc. of IEEE Conference on Computer Vision and Pattern
  Recognition (CVPR)}, pages 427--436, 2015.

\bibitem[Prest et~al.(2012)Prest, Leistner, Civera, Schmid, and
  Ferrari]{Prest:2012}
A.~Prest, C.~Leistner, J.~Civera, C.~Schmid, and V.~Ferrari.
\newblock Learning object class detectors from weakly annotated video.
\newblock In \emph{Proc. of the IEEE Conference on Computer Vision and Pattern
  Recognition (CVPR)}, pages 3282--3289, 2012.

\bibitem[Quian~Quiroga(2012)]{Quiroga:2012}
R.~Quian~Quiroga.
\newblock Concept cells: the building blocks of declarative memory functions.
\newblock \emph{Nature Reviews Neuroscience}, 13\penalty0 (8):\penalty0
  587--597, 2012.

\bibitem[Quian~Quiroga et~al.(2005)Quian~Quiroga, Reddy, Kreiman, Koch, and
  Fried]{Quiroga:2005}
R.~Quian~Quiroga, L.~Reddy, G.~Kreiman, C.~Koch, and I.~Fried.
\newblock Invariant visual representation by single neurons in the human brain.
\newblock \emph{Nature}, 435\penalty0 (7045):\penalty0 1102--1107, 2005.

\bibitem[Rudin(1991)]{Rudin:1991}
W.~Rudin.
\newblock \emph{Functional Analysis}.
\newblock International Series in Pure and Applied Mathematics. McGraw-Hill,
  1991.

\bibitem[Russakovsky et~al.(2014)Russakovsky, Deng, Su, Krause, Satheesh, Ma,
  Huang, Karpathy, Khosla, Bernstein, Berg, and Fei-Fei]{Russ:2015}
O.~Russakovsky, J.~Deng, H.~Su, J.~Krause, S.~Satheesh, S.~Ma, Z.~Huang,
  A.~Karpathy, A.~Khosla, M.~Bernstein, A.~C. Berg, and L.~Fei-Fei.
\newblock Imagenet large scale visual recognition challenge.
\newblock \emph{Int. J. Comput. Vis.}, pages 1--42, 2014.
\newblock DOI:10.1007/s11263-015-0816-y.

\bibitem[Scardapane and Wang(2017)]{scardapane2017randomness}
S.~Scardapane and D.~Wang.
\newblock Randomness in neural networks: an overview.
\newblock \emph{Wiley Interdisciplinary Reviews: Data Mining and Knowledge
  Discovery}, 7\penalty0 (2):\penalty0 e1200, 2017.

\bibitem[Schaefer(1999)]{Schaefer1999}
H.~Schaefer.
\newblock \emph{Topological Vector Spaces}.
\newblock Springer, New York, 1999.

\bibitem[Simonyan and Zisserman(2015)]{Simonyan:2015}
K.~Simonyan and A.~Zisserman.
\newblock Very deep convolutional networks for large-scale image recognition.
\newblock In \emph{International Conference on Learning Representations}, 2015.
\newblock arXiv:1409.1556.

\bibitem[Szegedy et~al.(2014)Szegedy, Zaremba, Sutskever, Bruna, Erhan,
  Goodfellow, and Fergus]{ICLR:2014:Szegedy}
C.~Szegedy, W.~Zaremba, I.~Sutskever, J.~Bruna, D.~Erhan, I.~J. Goodfellow, and
  R.~Fergus.
\newblock Intriguing properties of neural networks.
\newblock In \emph{Proc. of International Conference on Learining
  Representations (ICLR)}, 2014.
\newblock https://arxiv.org/abs/1312.6199.

\bibitem[Team(2016)]{Deeplearning4j}
D.~D. Team.
\newblock Deeplearning4j: Open-source distributed deep learning for the jvm.
\newblock Apache Software Foundation License 2.0. http://deeplearning4j.org,
  2016.

\bibitem[Vapnik(2000)]{Vapnik2000}
V.~Vapnik.
\newblock \emph{The Nature of Statistical Learning Theory}.
\newblock Springer-Verlag, 2000.

\bibitem[Vapnik and Chapelle(2000)]{Vapnik:2000}
V.~Vapnik and O.~Chapelle.
\newblock Bounds on error expectation for support vector machines.
\newblock \emph{Neural Computation}, 12\penalty0 (9):\penalty0 2013--2036,
  2000.

\bibitem[Viskontas et~al.(2009)Viskontas, Quian~Quiroga, and
  Fried]{Quiroga:2009}
I.~Viskontas, R.~Quian~Quiroga, and I.~Fried.
\newblock Human medial temporal lobe neurons respond preferentially to
  personally relevant images.
\newblock \emph{Proc. Nat. Acad. Sci.}, 106\penalty0 (50):\penalty0
  21329--21334, 2009.

\bibitem[Wang(2016)]{Wang:2016}
D.~Wang.
\newblock Editorial: Randomized algorithms for training neural networks.
\newblock \emph{Information Sciences}, 364--365:\penalty0 126--128, 2016.

\bibitem[Zheng et~al.(2016)Zheng, Song, Leung, and Goodfellow]{Zheng:2016}
S.~Zheng, Y.~Song, T.~Leung, and I.~Goodfellow.
\newblock Improving the robustness of deep neural networks via stability
  training.
\newblock In \emph{Proc. of the IEEE Conference on Computer Vision and Pattern
  Recognition (CVPR)}, 2016.
\newblock https://arxiv.org/abs/1604.04326.

\end{thebibliography}

%

\end{document}